\documentclass{article}

\PassOptionsToPackage{numbers, compress}{natbib}

\usepackage[preprint]{neurips_2026}

\usepackage[utf8]{inputenc}
\usepackage[T1]{fontenc}
\usepackage{hyperref}
\usepackage{url}
\usepackage{booktabs}
\usepackage{amsfonts}
\usepackage{amsmath}
\usepackage{amssymb}
\usepackage{nicefrac}
\usepackage{microtype}
\usepackage{xcolor}
\usepackage{graphicx}
\usepackage{multirow}
\usepackage{subcaption}
\usepackage{enumitem}
\usepackage{tikz}
\usetikzlibrary{positioning, decorations.pathreplacing, calc, arrows.meta, shapes.geometric, shapes.misc}
\usepackage{pgfplots}
\pgfplotsset{compat=1.18}

\newcommand{\pp}{\,\text{pp}}

\title{The Compliance Trap: Diagnosing How AI Agents Consume Conflicting Memory}

\author{
\textbf{Yixiong Chen}$^{1}$,
\textbf{Xinyi Bai}$^{2}$,
\textbf{Alan Yuille}$^{1}$ \\
$^{1}$Johns Hopkins University \quad
$^{2}$Cornell University \\
\texttt{ychen646@jh.edu}
}

\begin{document}

\maketitle

\begin{abstract}
Memory is becoming a core component of long-horizon AI agents, allowing agents to reuse past experience when operating web browsers, software tools, and other interactive environments. Existing work mostly treats memory as a \emph{supply} problem, asking what experience to write, how to store it, and which entry to retrieve for the next task. Yet we still lack a clear account of how models consume retrieved memory across a multi-step action trajectory. This \emph{consumption} process matters because it determines not only what memories should be retrieved, but also what models and control policies are needed to use them safely.
To diagnose this process, we propose \textbf{E}ntry--\textbf{P}ropagation--\textbf{R}ecovery (E-P-R), a trajectory-level framework that asks where memory first changes an action, whether that change carries forward, and whether the agent can recover after leaving a correct path.
We instantiate E-P-R on WebArena and on MemTrapBench, a controlled benchmark we build to isolate these phases.
We find that the main failure often begins at entry: agents adopt conflicting memory at the first exposed decision point even when it is task-wrong. Repeated exposure then amplifies this early error, while recovery after divergence is weak. Together, these effects create a \textbf{compliance trap}: across models, conflicting memory induces similar compliance rates, but once agents comply, their success rates collapse to a low floor. Stronger agents therefore suffer larger absolute damage because each compliance event erases more baseline capability.
These results suggest that memory-augmented agents should be evaluated not only by retrieval quality or final success rate, but by how they consume memory throughout the trajectory.
\end{abstract}

\section{Introduction}
\label{sec:intro}

Long-horizon agents are increasingly used to operate web browsers, file systems, software tools, and operating systems~\citep{wang2024survey, xi2023agent}. As these tasks span many observations and actions, agents need a way to reuse past experience instead of solving each task from scratch. Memory has become a common mechanism for this purpose. Trajectories from past episodes can be summarized, stored, and retrieved at the start of a new task to guide later decisions~\citep{shinn2023reflexion, zhao2024expel, wang2025awm, liu2025webcoach, packer2024memgpt, xu2025amem, zhong2024memorybank}. This makes memory a central component of practical agent systems, especially when the task is too long to fit all useful experience into the model context or too costly to reason through repeatedly.

Most existing work studies agent memory from the \emph{supply} side. 
Systems such as Reflexion~\citep{shinn2023reflexion}, ExpeL~\citep{zhao2024expel}, AWM~\citep{wang2025awm}, WebCoach~\citep{liu2025webcoach}, MemGPT~\citep{packer2024memgpt}, A-MEM~\citep{xu2025amem}, MemoryBank~\citep{zhong2024memorybank}, and RAP~\citep{kagaya2024rap} differ in how they write, retrieve, or summarize experience, but they share an implicit assumption: once a memory is placed in context, the agent policy will decide how to use it.
However, the agent that consumes memory is still an imperfect language model policy, and retrieval does not determine how the model will use the retrieved content. The useful memory can be ignored, followed too literally, or applied again after the state has changed. As a result, a memory that looks plausible to a retriever can be dangerous if the model adopts it at the wrong point or fails to recover from it later. Prior work has observed related failures in experience following~\citep{xiong2025experience}, but we still lack a trajectory-level account of where memory first changes behavior, how far that change carries forward, and whether the agent can undo it.

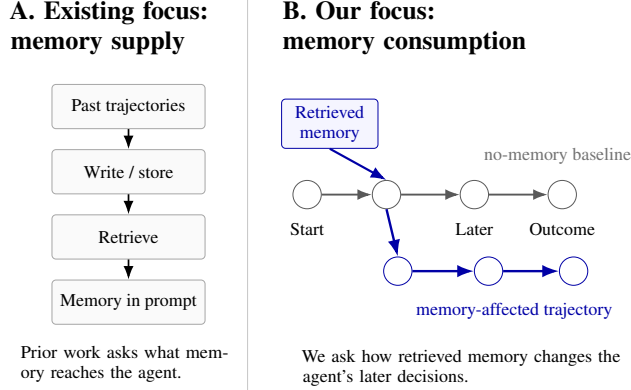
\begin{figure*}[t]
\centering
\resizebox{0.6\textwidth}{!}{%
\begin{tikzpicture}[
    x=1cm, y=1cm,
    font=\small,
    >=Latex,
    panelsep/.style={draw=gray!45, line width=0.5pt},
    flowbox/.style={draw=gray!70, rounded corners=2pt, minimum width=2.7cm, minimum height=0.78cm, align=center, fill=gray!4},
    smallcallout/.style={draw=black!70, rounded corners=2pt, fill=white, inner sep=2pt, font=\small},
    greenline/.style={draw=green!60!black, line width=1.1pt},
    redline/.style={draw=red!75!black, line width=1.1pt},
    grayline/.style={draw=black!65, line width=0.95pt},
    nodeneutral/.style={circle, draw=black!65, fill=white, minimum size=5.0mm, inner sep=0pt},
    nodegood/.style={circle, draw=green!60!black, fill=white, minimum size=5.0mm, inner sep=0pt},
    nodebad/.style={circle, draw=red!75!black, fill=white, minimum size=5.0mm, inner sep=0pt}
]

\def\H{7.5}
\def\xA{0.0}
\def\xB{4.8}
\def\xC{12.5}

\draw[panelsep] (\xB-0.5,0.8) -- (\xB-0.5,\H+0.5);

\node[anchor=west, align=left, text width=4.2cm,
      font=\bfseries\fontsize{13}{15}\selectfont] at (\xA,7.35)
{A. Existing focus:\\memory supply};

\node[flowbox, anchor=north] (a1) at (\xA+2.2,6.35) {Past trajectories};
\node[flowbox, anchor=north] (a2) at (\xA+2.2,5.20) {Write / store};
\node[flowbox, anchor=north] (a3) at (\xA+2.2,4.05) {Retrieve};
\node[flowbox, anchor=north] (a4) at (\xA+2.2,2.90) {Memory in prompt};

\draw[-Latex, line width=0.85pt] (a1.south) -- (a2.north);
\draw[-Latex, line width=0.85pt] (a2.south) -- (a3.north);
\draw[-Latex, line width=0.85pt] (a3.south) -- (a4.north);

\node[anchor=north, align=left, text width=3.8cm] at (\xA+2.2,1.85)
{Prior work asks what memory reaches the agent.};

\node[anchor=west, align=left, text width=5.7cm,
      font=\bfseries\fontsize{13}{15}\selectfont] at (\xB,7.35)
{B. Our focus:\\memory consumption};

\begin{scope}[yshift=0.5cm]

\node[draw=blue!65!black, rounded corners=2pt, fill=blue!4,
      text=blue!65!black, minimum width=1.70cm, minimum height=0.78cm,
      align=center] (mem) at (\xB+0.95,5.15)
{Retrieved\\memory};

\foreach \i/\x/\lab/\xshift in {
    1/0.55/Start/0,
    2/1.95/ /+0.48,
    3/3.50/Later/0,
    4/5.05/Outcome/0
} {
    \node[nodeneutral] (b\i) at (\xB+\x,3.90) {};
    \node[anchor=north, font=\small] at ($(b\i.south)+(\xshift,-0.12)$) {\lab};
}
\foreach \i/\j in {1/2,2/3,3/4} {
    \draw[grayline,-Latex] (b\i) -- (b\j);
}

\node[anchor=west, font=\small, text=black!55] at (\xB+3.55,4.58)
{no-memory baseline};

\draw[blue!65!black, line width=1.1pt, -Latex] (mem.south) -- (b2.north);

\foreach \i/\x in {1/2.15,2/3.75,3/5.25} {
    \node[circle, draw=blue!65!black, fill=white, minimum size=5.0mm, inner sep=0pt] (m\i) at (\xB+\x,2.55) {};
}
\draw[blue!65!black, line width=1.2pt, -Latex] (b2.south) -- (m1.north);
\foreach \i/\j in {1/2,2/3} {
    \draw[blue!65!black, line width=1.2pt, -Latex] (m\i) -- (m\j);
}

\node[anchor=west, font=\small, text=blue!65!black] at (\xB+2.35,1.85)
{memory-affected trajectory};

\node[anchor=west, align=left, text width=5.6cm] at (\xB+0.35,0.85)
{We ask how retrieved memory changes the agent's later decisions.};

\end{scope}

\end{tikzpicture}%
}
\caption{\textbf{Motivation and Overview.} (A) Prior work studies the memory supply side. (B) We analyze memory consumption: how trajectories change with/without memory injection. 
}
\label{fig:intro_teaser}
\vspace{-0.3cm}
\end{figure*}

Figure~\ref{fig:intro_teaser} illustrates this shift in perspective. Existing memory systems mainly control the supply pipeline: past trajectories are written, stored, retrieved, and inserted into the prompt. Our focus begins after this point. Once a memory enters the agent context, the agent must still decide whether to follow it, ignore it, repeat it after the state has changed, or correct course after it has caused a wrong action.
We therefore study memory as a consumption problem. \emph{We ask how long-horizon agents consume retrieved memory across a multi-step trajectory, and which stage of the trajectory determines whether memory helps or hurts.} To answer this, we introduce \textbf{E}ntry--\textbf{P}ropagation--\textbf{R}ecovery (E-P-R), a framework that decomposes memory consumption into three phases.
Entry measures whether retrieved memory changes the agent's action at the first decision point where it is exposed. Propagation measures whether this initial change continues to shape later actions under repeated or continued memory exposure. Recovery measures whether the agent can return to a task-correct trajectory after memory-induced divergence.

We apply this framework on two complementary testbeds. 
On WebArena, we hold memory content fixed and vary only the injection schedule. Injection on different stage (early/persistent/late) probes how memory enters the trajectory, amplifies the effect, and changes existing decision. These schedule interventions give coarse trajectory-level probes of E-P-R.
Across five models from Qwen3.5, Gemma4, and Gemini-3 families, the picture is consistent: wrong memory enters early, repeated exposure amplifies, and the damage concentrates on trajectories that follow the conflicting memory; once this happens, it is hard to recover. We call this regularity the \textbf{compliance trap}.
To test whether this pattern is specific to the WebArena, we then introduce MemTrapBench, a controlled benchmark for external validation.
By separately measuring uptake, persistence, and correction, MemTrapBench lets us test whether first adoption of conflicting memory is the main driver of final failure under a controlled trap structure.
Although E-P-R applies to helpful, conflicting, and cross-task control memory, our main analysis focuses on conflicting memory because it exposes whether the agent can reject task-wrong experience rather than merely follow retrieved text.

This paper makes three contributions. 
\begin{itemize}
    \item First, we formulate memory consumption as a trajectory-level problem and introduce E-P-R as a diagnostic framework for measuring entry, propagation, and recovery.
    \item Second, we identify the compliance trap as a concrete failure mode of memory-augmented agents, where similar compliance rates lead to larger absolute damage for stronger models.
    \item Third, we introduce MemTrapBench together with a paired WebArena protocol that separates full-benchmark average effects from mechanism diagnosis, showing that final success rate alone can hide where memory-induced failures arise.
\end{itemize}

\section{Related Work}
\label{sec:related}

\paragraph{Agent memory systems.}
Generative Agents~\citep{park2023generative} and Voyager~\citep{wang2023voyager} introduced persistent textual memory for open-ended interactive environments. Subsequent systems largely study memory from the supply side: Reflexion~\citep{shinn2023reflexion} writes failure reflections back into the prompt, ExpeL~\citep{zhao2024expel} distills insights across trajectories, Agent Workflow Memory~\citep[AWM;][]{wang2025awm} compiles reusable workflows with strong gains on WebArena, and WebCoach~\citep{liu2025webcoach} trains a coaching model on top of retrieved experience. A parallel line of work focuses on memory architecture rather than content: MemGPT~\citep{packer2024memgpt} treats large language models (LLMs) as operating systems and pages information between fast and slow memory tiers; MemoryBank~\citep{zhong2024memorybank} adds long-term semantic memory with forgetting curves; A-MEM~\citep{xu2025amem} structures memory as a self-organizing graph of agentic notes; RAP~\citep{kagaya2024rap} couples retrieval with planning for multimodal LLM agents; and Mem0~\citep{chhikara2025mem0} packages production-grade long-term memory for deployed assistants. Most closely related, \citet{xiong2025experience} analyze experience-following behavior in memory management systems. Our focus is complementary: we study how the agent policy consumes a given memory once it is in context.

\paragraph{Reasoning, self-correction, and recovery in language agents.}
A growing body of work asks whether language models can reason about and correct their own actions. Chain-of-thought prompting~\citep{wei2022cot}, self-consistency~\citep{wang2023selfconsistency}, and tree search~\citep{yao2023tot} extend single-step prediction with explicit deliberation, while ReAct~\citep{yao2023react} interleaves reasoning with environment interaction so that each action is conditioned on a verbalised plan. Self-Refine~\citep{madaan2023selfrefine}, Reflexion~\citep{shinn2023reflexion}, and critique-style training~\citep{bai2022constitutional, ouyang2022instructgpt} let models revise their outputs from feedback. Two recent surveys, however, document the limits of these approaches: \citet{huang2024selfcorrect} show that LLMs cannot reliably self-correct reasoning errors without an external signal, and \citet{kamoi2024selfcorrection} quantify the conditions under which self-correction helps or hurts. Our Recovery diagnostic targets whether the agent can leave a memory-induced trajectory after it has already taken several actions on the path. Empirically (Appendix~\ref{app:gate_annotation}), trajectory-level recovery exists but does not translate to task-level recovery.

\paragraph{Trajectory-level agent evaluation and failure diagnosis.}
WebArena~\citep{zhou2024webarena}, VisualWebArena~\citep{koh2024visualwebarena}, OSWorld~\citep{xie2024osworld}, MiniWoB++~\citep{shi2017miniwob}, WorkArena~\citep{drouin2024workarena}, Mind2Web~\citep{deng2023mind2web}, WebShop~\citep{yao2022webshop}, WebVoyager~\citep{he2024webvoyager}, AgentBench~\citep{liu2024agentbench}, AgentBoard~\citep{ma2024agentboard}, and TheAgentCompany~\citep{xu2024theagentcompany} evaluate agents on multi-step interactive tasks, usually through final task success or per-step progress. A separate strand develops the visual grounding side of these agents: SeeClick~\citep{cheng2024seeclick}, CogAgent~\citep{hong2024cogagent}, SeeAct~\citep{zheng2024seeact}, and Set-of-Mark prompting~\citep{yang2023setofmark} show that linking instructions to concrete UI elements can dominate raw planning quality, an observation that motivates our explicit Grounding gate (Section~\ref{sec:results}). Earlier work on browser-assisted question answering, including WebGPT~\citep{nakano2021webgpt} and the BrowserGym ecosystem~\citep{drouin2024browsergym}, established the action interface that all of these benchmarks now share. Recent diagnostic benchmarks such as WebSuite~\citep{li2024websuite} go beyond binary success by attributing failures to specific web-action types, and AgentBoard~\citep{ma2024agentboard} reports fine-grained progress rates. These analyses identify where agents fail in the environment. Our analysis instead asks how an injected memory changes the trajectory.

\paragraph{Context utilization and misleading context.}
A separate line of work studies how language models use information provided in context. Lost in the Middle~\citep{liu2024lost} shows that position within a long context affects whether relevant information is used. CUB~\citep{hagstrom2025cub} benchmarks context-utilization methods under different context conditions. Work on parametric-context conflict studies when models follow retrieved evidence versus internal knowledge~\citep{xu2024knowledge, lewis2020rag}, while sycophancy work shows that models can agree with misleading user-provided pressure~\citep{perez2023sycophancy, sharma2024sycophancy, wei2024sycophancy}. These studies mainly evaluate static responses, while we study the analogous problem in interactive agents.

\section{Method}
\label{sec:method}

We study memory consumption by intervening on two variables: \emph{what} the memory says and \emph{when} the agent can see it. A memory is a short textual passage inserted into the agent context during an episode. Its content is one of three types. A helpful memory gives task-relevant guidance, a conflicting memory gives plausible but task-wrong guidance, and a same-website cross-task control memory provides a content control. Its schedule specifies the decision steps at which the memory is available: early (only the first decision point), persistent (every decision point), or late (only after the first few actions have already been taken).

By fixing the memory passage and changing only its availability over time, we test the agent policy's use of memory rather than the quality of retrieval. This setting follows the common design of external textual memories used in systems such as Reflexion~\citep{shinn2023reflexion}, ExpeL~\citep{zhao2024expel}, and AWM~\citep{wang2025awm}.

\subsection{Entry--Propagation--Recovery diagnostics}
\label{sec:epr}

\begin{figure}[t]
\centering
\resizebox{0.98\linewidth}{!}{%

\begin{tikzpicture}[x=1cm, y=1cm, line cap=round, line join=round, >=Latex]

\definecolor{confred}{RGB}{150,40,55}
\definecolor{helpblue}{RGB}{36,92,150}
\definecolor{okgreen}{RGB}{46,125,50}
\definecolor{midgray}{RGB}{120,120,120}
\definecolor{lightpanel}{RGB}{245,246,248}
\definecolor{xred}{RGB}{198,40,40}

\sffamily


\begin{scope}
  \draw[confred, line width=1pt, fill=white]
    (0.0,2.10) -- (1.0,2.10) -- (1.0,2.95) -- (0.75,3.20) -- (0.0,3.20) -- cycle;
  \draw[confred, line width=0.8pt, fill=confred!15]
    (1.0,2.95) -- (0.75,2.95) -- (0.75,3.20) -- cycle;
  \draw[xred, line width=1.6pt] (0.35,2.45) -- (0.65,2.75);
  \draw[xred, line width=1.6pt] (0.65,2.45) -- (0.35,2.75);
\end{scope}
\node[font=\scriptsize, align=center, text width=2cm] at (0.5,1.78) {Conflicting\\memory};

\begin{scope}
  \draw[helpblue, line width=1pt, fill=white]
    (0.0,0.10) -- (1.0,0.10) -- (1.0,0.95) -- (0.75,1.20) -- (0.0,1.20) -- cycle;
  \draw[helpblue, line width=0.8pt, fill=helpblue!15]
    (1.0,0.95) -- (0.75,0.95) -- (0.75,1.20) -- cycle;
  \draw[helpblue, line width=1.6pt]
    (0.30,0.62) -- (0.45,0.46) -- (0.72,0.82);
\end{scope}
\node[font=\scriptsize, align=center, text width=2cm] at (0.5,-0.22) {Helpful\\memory};

\draw[fill=black!20, draw=black, line width=1pt, rounded corners=2pt]
  (1.90,1.10) rectangle (2.85,1.65);
\draw[fill=black!20, draw=black, line width=1pt, rounded corners=2pt]
  (1.95,1.70) rectangle (2.80,2.20);
\draw[black, line width=0.8pt] (2.38,2.20) -- (2.38,2.40);
\fill[black] (2.38,2.44) circle (0.05);
\fill[black] (2.18,1.95) circle (0.055);
\fill[black] (2.58,1.95) circle (0.055);
\node[font=\scriptsize] at (2.38,0.82) {agent};

\draw[->, line width=0.8pt] (1.05,2.55) -- (1.90,1.98);
\draw[->, line width=0.8pt] (1.05,0.62) -- (1.90,1.45);


\fill[lightpanel, rounded corners=3pt] (3.35,0.20) rectangle (5.95,3.88);
\fill[lightpanel, rounded corners=3pt] (6.25,0.20) rectangle (10.90,3.88);
\fill[lightpanel, rounded corners=3pt] (11.20,0.20) rectangle (15.25,3.88);

\draw[black!15, rounded corners=3pt] (3.35,0.20) rectangle (5.95,3.88);
\draw[black!15, rounded corners=3pt] (6.25,0.20) rectangle (10.90,3.88);
\draw[black!15, rounded corners=3pt] (11.20,0.20) rectangle (15.25,3.88);



\node[font=\Large] at (4.65,4.76) {\textbf{E}\;\;Entry};
\node[font=\scriptsize\itshape, text=midgray, align=center, text width=2.9cm] at (4.65,4.18)
  {does memory change\\the action?};

\node[font=\Large] at (8.58,4.76) {\textbf{P}\;\;Propagation};
\node[font=\scriptsize\itshape, text=midgray, align=center, text width=3.6cm] at (8.58,4.18)
  {does the change persist\\under later steps?};

\node[font=\Large] at (13.22,4.76) {\textbf{R}\;\;Recovery};
\node[font=\scriptsize\itshape, text=midgray, align=center, text width=3.8cm] at (13.22,4.18)
  {can harmful divergence\\be corrected?};


\draw[midgray, dashed, line width=1.4pt] (3.10,2.05) -- (14.95,2.05);

\draw[->, line width=0.9pt] (2.80,2.05) -- (3.10,2.05);

\node[font=\scriptsize, text=midgray, fill=white, inner sep=1.2pt] at (11.65,1.83) {no-memory baseline};

\draw[black!35, densely dotted, line width=0.8pt] (4.65,0.62) -- (4.65,3.42);

\draw[helpblue, line width=2.3pt]
  (3.10,2.05) -- (4.78,2.05)
  .. controls (5.20,2.05) and (5.38,1.16) .. (6.05,1.16)
  -- (14.55,1.16);
\draw[helpblue, line width=2.3pt, ->] (14.55,1.16) -- (14.95,1.16);

\draw[confred, line width=2.3pt]
  (3.10,2.05) -- (4.78,2.05)
  .. controls (5.20,2.05) and (5.38,3.00) .. (6.05,3.00)
  -- (11.35,3.00);

\draw[confred, line width=2.3pt] (11.35,3.00) -- (14.55,3.00);
\draw[confred, line width=2.3pt, ->] (14.55,3.00) -- (14.95,3.00);

\draw[confred, line width=1.5pt, ->]
  (11.35,3.00) .. controls (12.10,3.00) and (12.50,2.10) .. (13.70,2.05);



\node[font=\scriptsize, text=confred, align=center, text width=3.5cm] at (8.65,2.52)
  {harmful divergence can \\persist across later actions};

\node[font=\scriptsize, text=helpblue, align=center, text width=3.2cm] at (8.55,1.58)
  {grounded helpful memory\\improves the trajectory};

\node[font=\scriptsize, text=confred, align=center] at (13.15,2.52)
  {correct};


\draw[okgreen, line width=3pt] (15.70,1.18) -- (15.86,1.00) -- (16.18,1.36);
\node[font=\small, anchor=west] at (16.42,1.18) {success};

\draw[xred, line width=3pt] (15.68,2.75) -- (16.10,3.25);
\draw[xred, line width=3pt] (16.10,2.75) -- (15.68,3.25);
\node[font=\small, anchor=west] at (16.42,3.00) {failure};


\draw[rounded corners=2pt, fill=white, draw=black!20] (3.5,0.58) rectangle (5.9,1.04);
\node[font=\scriptsize, align=center] at (4.72,0.81)
  {first changed action};

\draw[rounded corners=2pt, fill=white, draw=black!20] (6.95,0.58) rectangle (10.05,1.04);
\node[font=\scriptsize, align=center] at (8.50,0.81)
  {schedule effect};

\draw[rounded corners=2pt, fill=white, draw=black!20] (12.10,0.58) rectangle (14.20,1.04);
\node[font=\scriptsize, align=center] at (13.15,0.81)
  {self-correction};


\draw[midgray, dashed, line width=1.2pt] (3.55,-0.40) -- (4.35,-0.40);
\node[font=\scriptsize, anchor=west] at (4.50,-0.40) {no-memory baseline};

\draw[helpblue, line width=2.2pt] (7.15,-0.40) -- (7.95,-0.40);
\node[font=\scriptsize, anchor=west] at (8.10,-0.40) {helpful-memory trajectory};

\draw[confred, line width=2.2pt] (11.45,-0.40) -- (12.25,-0.40);
\node[font=\scriptsize, anchor=west] at (12.40,-0.40) {conflicting-memory trajectory};

\end{tikzpicture}

}
\caption{\textbf{Entry--Propagation--Recovery diagnostic framework.} \textbf{Entry}: does retrieved memory change the agent's action? \textbf{Propagation}: does the change persist under later observations? \textbf{Recovery}: can the agent correct a harmful deviation? Helpful memory improves trajectories when grounded; conflicting memory harms when adopted and not recovered.\label{fig:epr_framework}}
\vspace{-0.25cm}
\end{figure}
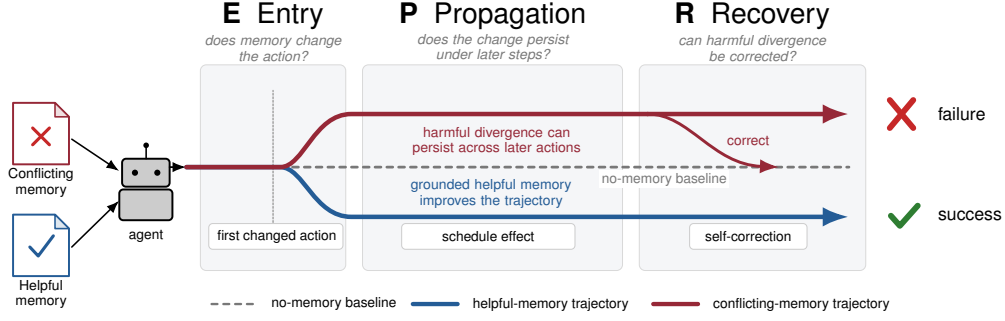

We treat memory consumption as a trajectory-level process rather than a single retrieval event. A retrieved memory first competes with the no-memory policy at the next decision point; if it changes the agent's action, the trajectory branches away from the no-memory baseline; the branch can then persist through later observations or be corrected. Figure~\ref{fig:epr_framework} summarizes the three stages this defines: \textbf{Entry} asks whether memory changes the first action it can affect, \textbf{Propagation} asks whether that change carries forward under continued memory exposure, and \textbf{Recovery} asks whether the agent returns to a task-correct trajectory after divergence. The three stages together account for the full path from a memory passage to a final outcome, so a metric defined on any one of them is a partial proxy for the others.

We operationalize E-P-R through paired interventions. For each (task, model) cell we run one no-memory trajectory and one trajectory per (memory content, schedule) combination from the same environment seed; primary results are reported as the success-rate (SR) difference against the paired no-memory trace. Three injection schedules each probe one stage: \textbf{early} injection (memory visible only at step~1) targets Entry; \textbf{persistent} injection (memory visible at every step) reads off Propagation by amplifying or dampening the early effect; \textbf{late} injection (memory withheld until step~3) probes whether an already-formed action history resists new guidance, complementary to Recovery. Recovery itself is most directly read off MemTrapBench (Section~\ref{sec:design}), whose controlled task structure fixes a known correct path and a known trap action so trajectory-level realignment after a wrong action is observable~\citep{huang2024selfcorrect, kamoi2024selfcorrection}.


\subsection{Compliance under conflicting memory}
\label{sec:compliance_metric}

For conflicting memory, action divergence alone is not enough: the agent may differ from the no-memory trajectory without actually following the injected recommendation. We therefore measure compliance at the first memory-exposed decision step using the \textbf{Recommendation Compliance Rate (RCR)}: an LLM judge classifies whether the conflicting step-1 action plausibly executes the first primitive of the memory's recommendation, with the paired no-memory step-1 action as joint evidence (rubric in Appendix~\ref{app:edr_dpc}). Let $\mathcal{C}^{\mathrm{RCR}}$ denote the compliant subset (judge=YES). The cost of compliance is then the \emph{Damage Per Compliance} (DPC):
\[
\mathrm{DPC}(\mathcal{C}^{\mathrm{RCR}})=
\frac{1}{|\mathcal{C}^{\mathrm{RCR}}|}\sum_{i\in \mathcal{C}^{\mathrm{RCR}}}(y_i^{0}-y_i^{c}),
\]
where $y_i^{0}$ and $y_i^{c}$ are paired no-memory and conflicting-memory success on task $i$. Pairing on $\mathcal{C}^{\mathrm{RCR}}$ avoids comparing an all-task no-memory baseline to compliance success on a behavior-selected subset. DPC separates the \emph{rate} at which the agent commits to a wrong recommendation (RCR) from the \emph{conditional cost} of that commitment. The separation is essential because RCR turns out to be approximately constant across models (Section~\ref{sec:compliance_trap}, $\sim 65\%$), so unconditional success drops mostly reflect how far compliant trajectories fall rather than how often the agent complies. RCR and DPC together are the primary readout of the trap in Tables~\ref{tab:compliance} and~\ref{tab:memtrapbench}; a coarser action-divergence subset is reported as a robustness check in Appendix~\ref{app:edr_dpc}.

\subsection{Testbeds and protocol}
\label{sec:design}

\paragraph{Models.} We evaluate Qwen3.5-9B, Qwen3.5-27B, Gemma-4-E4B-it, Gemma-4-26B-A4B-it, and Gemini-3-Flash on both testbeds. The agent observes the page through the BrowserGym accessibility tree (AXTree)~\citep{drouin2024browsergym} and decodes greedily with vLLM, so two trajectories on the same task and seed differ only when memory injection differs.

\paragraph{Memories.} A memory is a short text passage placed in the agent context. All memories follow a single \texttt{DO}/\texttt{DON'T} template, and helpful and conflicting variants are matched in length, word count, and structural elements. Conflicting variants are plausible but task-wrong: they reference real UI elements on the same site and are written so that following them produces a well-formed but incorrect trajectory rather than a parse failure. WebArena memories are hand-authored. MemTrapBench memories are drafted by Claude Opus~4.7 from each task spec and reviewed by hand.

\paragraph{Statistics.} All paired contrasts use 10k-iteration sign-flip permutation tests; reported confidence intervals are paired bootstrap percentiles with 10k resamples. Full audits, inference settings, and the RCR rubric are in Appendix~\ref{app:reproducibility}.

\paragraph{WebArena.} \citep{zhou2024webarena} We use the 684 non-map tasks for the full-benchmark pre-check and a 77-task memory-sensitive subset (tasks whose outcome changes under early helpful or conflicting memory on Qwen3.5-9B or Qwen3.5-27B) for trajectory-level analysis.

\paragraph{MemTrapBench.} 231 long-horizon (median 15 steps) browser tasks on a shared HTML/JS skeleton served through BrowserGym. Each task pairs a verifiable goal with a plausible decoy and falls into one of four diagnostic families, each targeting one E-P-R gate by construction:
\begin{itemize}[leftmargin=1.4em,topsep=2pt,itemsep=1pt,parsep=0pt]
  \item \textsc{Decoy} (Entry): a prominent decoy sits beside the correct path.
  \item \textsc{Uptake} (Adoption): the alternate route is hidden behind an Advanced toggle.
  \item \textsc{Grounding} (Grounded$\mid$Adopted): a memory-referenced UI label is absent from the AXTree.
  \item \textsc{Override} (Recovery): memory tells the agent to ignore an explicit environment correction.
\end{itemize}
Together these decompose the helpful lift into a multiplicative gate chain $\Delta_{\text{helpful}}=P(\text{novel})\cdot P(\text{adopt})\cdot P(\text{ground}\mid\text{adopt})\cdot P(\text{lift}\mid\text{ground})$; the same gates govern conflicting damage. Tasks and memories are authored from per-family blueprints by Claude Opus~4.7 and verified by hand: each task is rolled out once under no-memory and once under persistent-conflicting during construction to confirm the correct path succeeds and the trap action lies on a wrong path. Templates, family counts, and audit are in Appendix~\ref{app:mtb_templates}.

\section{Results}
\label{sec:results}

We organize the results around the E-P-R chain. E-P-R is not a claim that retrieved memory is generally harmful. It treats memory as a high-leverage input whose effect depends on whether it enters the trajectory, remains grounded in the page state, and can be corrected when wrong. Our main contribution is a quantitative account of the unsafe direction: prior work on agent memory has focused almost entirely on raising the upside of helpful memory, while the failure mode in which a task-wrong memory enters the trajectory and resists correction has not been systematically characterised. We therefore analyze helpful and conflicting memory together: helpful memory is the positive control showing that the same gates control benefit when novelty and grounding are satisfied, while conflicting memory exposes the unsafe regime in which the same gates let a wrong recommendation control the trajectory.


\subsection{Full-WebArena pre-check motivates consumption-side diagnosis}
\label{sec:motivation}

Before any hand authoring or outcome-based task selection, we run a motivation check (Table~\ref{tab:full_webarena_retrieval_precheck}) on the full 684-task WebArena distribution with online-generated memories in the contrastive DO/DON'T format. The effect on Qwen3.5-9B and 27B are not consistent ($+1.7\pp$ vs $-1.9\pp$). After trying raising the relevance threshold to increase the memory quality, no performance gain is observed: damage worsens slightly on Qwen3.5-9B, and Qwen3.5-27B flips from a small gain to a small loss. This pre-check experiment shows that memory does not automatically bring benefit to tasks, especially when the task is complex and long-horizon. An in-depth exploration of how memory affects the agent's behavior is thus needed.

\begin{table}[t]
\caption{\textbf{Full-WebArena retrieval pre-check.} Online-generated memory across 684 non-map tasks. Memory based on the simple contrastive setting cannot uniformly improve the model performance with no memory, no matter the retrieved memory is gated or not.}
\label{tab:full_webarena_retrieval_precheck}
\centering
\small
\begin{tabular}{lcc}
\toprule
Model & Unconditional retrieval & Relevance-gated retrieval \\
\midrule
Qwen3.5-9B  & $-1.9\pp$ & $-2.3\pp$ \\
Qwen3.5-27B & $+1.7\pp$ & $-1.7\pp$ \\
\bottomrule
\end{tabular}
\end{table}

\subsection{Conflicting memory enters early and persistent exposure amplifies harm}
\label{sec:webarena_epr}

\begin{figure}[t]
  \centering
  \includegraphics[width=0.98\textwidth, trim=1.8cm 0 1.8cm 0, clip]{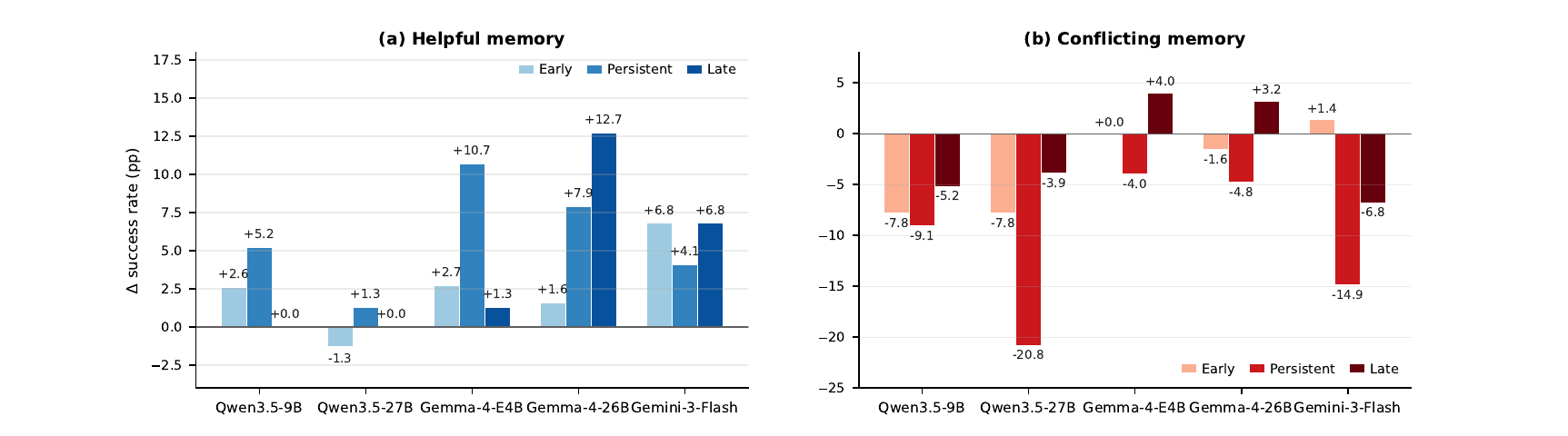}
  \caption{\textbf{WebArena schedule deltas on the 77-task subset (5 models, including Gemini-3-Flash, $n{=}74$).} (a) Helpful memory: modest gains, mostly persistent. (b) Persistent conflicting hurts every model; peaks on Qwen3.5-27B ($-20.8\pp$) and reaches $-14.9\pp$ on Gemini-3-Flash. Late conflicting is weak on WebArena's $\approx 7$-step median; MemTrapBench (Section~\ref{sec:memtrapbench}) restores late helpful. Cross-task control within $\pm 3\pp$ (Appendix~\ref{app:webarena_extended}).}
  \label{fig:schedule_bars}
\end{figure}

Figure~\ref{fig:schedule_bars} shows success-rate deltas under early, persistent, and late injection for helpful and conflicting memory across five models on the 77-task subset. Persistent conflicting memory is the only condition that hurts every model, with significant drops on Qwen3.5-27B and Gemini-3-Flash; the same direction holds for the smaller open-weight models. Persistent conflicting is also significantly worse than early conflicting on Qwen3.5-27B ($p=0.013$); the harm is not a single misclick but accumulates with repeated exposure, consistent with Propagation. Late injection has only a weak effect because the median WebArena trajectory is $\approx 7$ steps and the agent has already committed to an action history.

Helpful memory yields an asymmetrical effect on models. It only has modest improvement on the models that are sensitive to the conflicting memories (Qwen3.5-27B and Gemini-3-Flash), but helps more for the models that resist conflicting memories (Gemma-4-E4B and Gemma-4-26B-A4B). The asymmetry indicates that the behavior of models after memory injection is specific to the model family and size, putting more importance on the consumption-side exploration.

\subsection{The compliance trap: similar compliance, larger conditional damage}
\label{sec:compliance_trap}

We focus on the WebArena subset that all the memories' following effect can be verified by LLMs. Table~\ref{tab:compliance} reports the RCR and paired DPC on all five WebArena models. Three observations define the compliance trap. (i) RCR is high and approximately scale-independent (63--72\% across the five models). (ii) Conditional success after compliance collapses to a near-constant floor (17--31\%), well below the no-memory baseline on the same subset (23--49\%). (iii) Because higher-baseline models start higher but land on a similar low floor, paired DPC tracks baseline capability: from $-2.1\pp$ on Gemma-4-E4B (low baseline 23\%) to $-25.5\pp$ on Qwen3.5-27B (baseline 49\%). On WebArena $n{=}77$ the CIs exclude zero on the two highest-baseline models (Qwen3.5-27B and Gemini-3-Flash); the smaller open-weight models show the same direction with smaller magnitude. 

\begin{table}[t]
  \caption{\textbf{Paired compliance trap under persistent conflicting memory on WebArena 77.} $\mathcal{C}^{\mathrm{RCR}}$ indicates LLM-judge step-1 Recommendation Compliance subset. SR$^{0}_{\mathcal{C}^{\mathrm{RCR}}}$ and SR$^{c}_{\mathcal{C}^{\mathrm{RCR}}}$ are success rates for no memory / with memory. Paired DPC$=\mathrm{SR}^{0}_{\mathcal{C}^{\mathrm{RCR}}}-\mathrm{SR}^{c}_{\mathcal{C}^{\mathrm{RCR}}}$.}
  \label{tab:compliance}
  \centering
  \small
  \setlength{\tabcolsep}{5pt}
  \begin{tabular}{lcccccc}
    \toprule
    Model & $N$ & RCR & $|\mathcal{C}^{\mathrm{RCR}}|$ & SR$^{0}_{\mathcal{C}^{\mathrm{RCR}}}$ & SR$^{c}_{\mathcal{C}^{\mathrm{RCR}}}$ & Paired DPC [95\% CI] \\
    \midrule
    Gemma-4-E4B-it       & 75 & 63\% & 47 & 23.4\% & 21.3\% & \phantom{1}2.1\pp\, [$-$10.6, +14.9] \\
    Gemma-4-26B-it       & 67 & 63\% & 42 & 38.1\% & 30.9\% & \phantom{1}7.1\pp\, [$-$4.8, +19.1] \\
    Qwen3.5-9B  & 77 & 69\% & 53 & 26.4\% & 17.0\% & \phantom{1}9.4\pp\, [$-$3.8, +22.6] \\
    Gemini-3-Flash       & 75 & 72\% & 54 & 46.3\% & 27.8\% & \textbf{18.5\pp\, [+3.7, +33.3]} \\
    Qwen3.5-27B & 77 & 66\% & 51 & 49.0\% & 23.5\% & \textbf{25.5\pp\, [+11.8, +39.2]} \\
    \bottomrule
  \end{tabular}
\end{table}

\begin{figure}[t]
  \centering
  \includegraphics[width=0.92\textwidth]{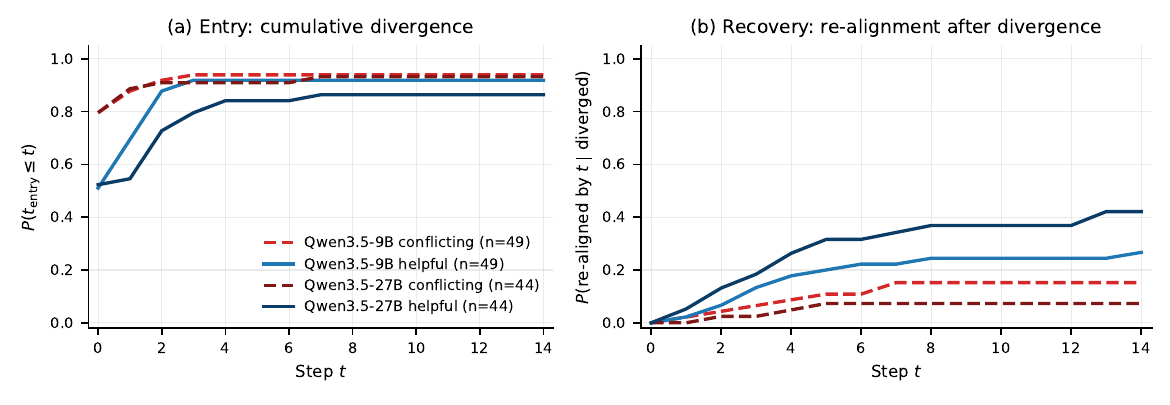}
  \caption{\textbf{Entry and recovery on WebArena under persistent injection.} (a) Memory enters early. $P(t_{\mathrm{entry}}\leq t)$ reaches 0.86--0.94 by step 7 on all models. (b) The recovery rates are low. Conditional on divergence, helpful trajectories re-align (27--42\%) more often than conflicting (7--15\%).}
\label{fig:epr_empirical}
\end{figure}

Figure~\ref{fig:epr_empirical} visualizes the trajectory-level signature of the trap on the two Qwen3.5 models. Panel (a) shows that the cumulative probability of first divergence reaches 0.86--0.94 by step 14 and is already high within the first few steps under both helpful and conflicting memory, so memory enters the trajectory early. Panel (b) shows that conditional on having diverged, helpful-memory trajectories re-align with the no-memory trajectory more often by step 14 (27--42\%) than conflicting-memory trajectories (7--15\%); the trap is therefore not only early adoption but harmful divergence that is less recoverable.

\subsection{MemTrapBench validates E-P-R without WebArena outcome selection}
\label{sec:memtrapbench}

The 77-task subset is selected on memory response, so its absolute effect sizes may be inflated. MemTrapBench (Section~\ref{sec:design}) addresses this by defining traps via construction rather than by downstream outcome, and by freezing helpful, conflicting, and cross-task control memories before any model evaluation.

Table~\ref{tab:memtrapbench} shows that the WebArena pattern is not a subset artifact. Persistent conflicting memory remains content-specifically harmful, while persistent cross-task control stays near zero. Damage grows with baseline capability: $-9.5\pp$ on Qwen3.5-9B, $-21.2\pp$ on Qwen3.5-27B, and $-26.0\pp$ on Gemini-3-Flash. The right two columns report paired DPC + 95\% CI on the trajectory-level divergence subset: \emph{all four models reach CI excluding zero}, addressing the WebArena sample-size limitation by replicating the trap on the larger 231-task pool with full statistical resolution on every model and across both open- and closed-weight families. Persistent helpful memory acts as a positive control on the same task pool: gains are large and uniform across models ($+25$ to $+31\pp$), confirming that our memory passages are usable when the gates allow them through. The cross-task control delta sits near zero, ruling out a context-length artifact. Late helpful memory nearly matches persistent helpful on the three open-weight models (Appendix~\ref{app:gate_annotation}), confirming that action-history inertia is a short-horizon effect.

\begin{table}[t]
\caption{\textbf{MemTrapBench persistent-injection deltas (231 long-horizon tasks).} Right two columns: paired DPC + 95\% CI on the action-divergence subset (step-1 action differs from no-memory); all four models exclude 0, $p<.001$.}
\label{tab:memtrapbench}
\centering
\small
\setlength{\tabcolsep}{4.5pt}
\begin{tabular}{lcccccc}
\toprule
Model & Base & Pers. Helpful & Pers. Conflict & Cross-task & Paired DPC & 95\% CI \\
\midrule
Qwen3.5-9B          & 10.8\% & \textbf{+25.5} & $-$9.5  & $-$2.6 & \textbf{+10.2}\pp & $[+5.4, +15.7]$ \\
Qwen3.5-27B         & 22.5\% & \textbf{+30.7} & $-$21.2 & $+$1.3 & \textbf{+26.6}\pp & $[+19.6, +33.5]$ \\
Gemma-4-26B         & 12.1\% & \textbf{+29.9} & $-$10.8 & $+$0.9 & \textbf{+14.1}\pp & $[+8.3, +19.9]$ \\
Gemini-3-Flash      & 32.9\% & \textbf{+29.4} & $\mathbf{-26.0}$ & $-$1.7 & \textbf{+28.6}\pp & $[+21.9, +35.2]$ \\
\bottomrule
\end{tabular}
\end{table}

Table~\ref{tab:mtb_diagnostic} breaks the per-diagnostic persistent injection deltas by family. \textsc{Decoy} produces the largest conflicting damage on every model because the wrong recommendation is groundable, immediately executable, and outcome-relevant; \textsc{Uptake} and \textsc{Override} produce substantial conflicting damage on the stronger models because the wrong recommendation is also reachable. The \textsc{Grounding} row shows the asymmetric helpful side of the same gate: when the conflicting recommendation references a UI element absent from the page, the helpful memory simultaneously carries the missing value the task requires, and helpful gains are the largest in the table ($+57.9$ to $+73.7\pp$). 

\begin{table}[t]
\caption{\textbf{Per-diagnostic persistent-injection deltas on MemTrapBench.} pH / pC are persistent helpful / conflicting deltas vs.\ no memory. \textsc{Decoy} (Entry) produces the largest conflicting damage; \textsc{Grounding} carries the largest helpful gain because the missing value is memory-borne. The conflicting-side reading on \textsc{Grounding} is constrained by a small baseline on the open-weight models ($\leq 1.8\%$); Gemini-3-Flash's higher baseline ($17.5\%$) reveals a clear $-15.8\pp$ pC. We test grounding as a causal gate via the $2\times 2$ factorial in Appendix~\ref{app:factorial} instead.}
\label{tab:mtb_diagnostic}
\centering
\footnotesize
\setlength{\tabcolsep}{3pt}
\begin{tabular}{l l cc cc cc cc}
\toprule
 & E-P-R Gate & \multicolumn{2}{c}{Qwen3.5-9B} & \multicolumn{2}{c}{Qwen3.5-27B} & \multicolumn{2}{c}{Gemma-4-26B} & \multicolumn{2}{c}{Gemini-3-Flash} \\
\cmidrule(lr){3-4}\cmidrule(lr){5-6}\cmidrule(lr){7-8}\cmidrule(lr){9-10}
Diagnostic & & pH & pC & pH & pC & pH & pC & pH & pC \\
\midrule
\textbf{\textsc{Decoy}}      & \textbf{Entry}              & $\mathbf{+14.5}$ & $\mathbf{-24.2}$ & $\mathbf{+16.1}$ & $\mathbf{-35.5}$ & $\mathbf{+19.4}$ & $\mathbf{-25.8}$ & $\mathbf{+12.9}$ & $\mathbf{-43.5}$ \\
\textsc{Uptake}             & Adoption                    & $+19.6$ & $\phantom{-}0.0$ & $+17.9$ & $-21.4$ & $+19.6$ & $-10.7$ & $+8.9$  & $-19.6$ \\
\textsc{Grounding}          & Grounded$\mid$Adopted       & $+66.7$ & $\phantom{-}0.0$ & $+73.7$ & $-1.8$  & $+57.9$ & $-1.8$  & $+68.4$ & $-15.8$ \\
\textsc{Override}           & Recovery                    & $+1.8$  & $-12.5$ & $+16.1$ & $-25.0$ & $+23.2$ & $-3.6$  & $+28.6$ & $-23.2$ \\
\bottomrule
\end{tabular}\\[2pt]
{\footnotesize No-memory baselines by diagnostic (Qwen3.5-9B / Qwen3.5-27B / Gemma-4-26B / Gemini-3-Flash): \textsc{Decoy} 24.2/35.5/25.8/53.2\%, \textsc{Uptake} 3.6/26.8/16.1/35.7\%, \textsc{Grounding} 0.0/1.8/1.8/17.5\%, \textsc{Override} 14.3/25.0/3.6/23.2\%.}
\end{table}

\section{Discussion}
\label{sec:discussion}

\paragraph{Memory consumption, not system-prompt injection.} A natural follow-up is whether the compliance trap is a general property of memory consumption or only appears when memory is delivered through the system prompt. We re-run MemTrapBench with the same memory text inserted at the bottom of each observation instead of in the system prompt, so the agent reads memory alongside the page rather than as an instruction. Table~\ref{tab:footer_main} shows that persistent-conflicting damage is unchanged within $\pm 0.5\pp$ on all three open-weight models. The trap is therefore a property of how the agent uses memory content, not of the channel through which memory arrives: whether memory enters as a system instruction, a tool output, or part of the page, the same trajectory dynamics apply.

\begin{table}[h]
\vspace{-0.3em}
\caption{\textbf{Persistent-conflicting damage is invariant to memory placement.} Same memory text, different position; $\Delta$ vs.\ no-memory baseline on MemTrapBench (231 long-horizon tasks).}
\label{tab:footer_main}
\centering
\small
\setlength{\tabcolsep}{10pt}
\begin{tabular}{lcc}
\toprule
Model & System prompt $\Delta$ & Observation footer $\Delta$ \\
\midrule
Qwen3.5-9B   & $-9.5\pp$  & $-10.0\pp$ \\
Qwen3.5-27B  & $-21.2\pp$ & $-20.8\pp$ \\
Gemma-4-26B  & $-10.8\pp$ & $-10.4\pp$ \\
\bottomrule
\end{tabular}
\vspace{-0.3em}
\end{table}

\paragraph{The trap requires a non-trivial trajectory length.} Across three benchmarks with different median trajectory lengths (Table~\ref{tab:horizon}), persistent-conflicting damage scales with horizon: short-horizon MiniWoB++ shows null effects, the trap emerges around 5--7 steps on WebArena, and stabilizes on the 15-step MemTrapBench. The trap is therefore a propagation phenomenon, not a single-step misclick, and is consistent with reports that LLM self-correction is most fragile precisely when the agent must integrate feedback over many steps~\citep{huang2024selfcorrect, kamoi2024selfcorrection}.

\begin{table}[h]
\vspace{-0.3em}
\caption{\textbf{Persistent conflicting damage scales with median trajectory length.}}
\label{tab:horizon}
\centering
\small
\setlength{\tabcolsep}{5pt}
\begin{tabular}{lcccc}
\toprule
Benchmark & Med.\ steps & Qwen3.5-9B & Qwen3.5-27B & Gemma-4-26B \\
\midrule
MiniWoB++ ($n{=}36$)         & 3  & $\phantom{+}0.0$ & $-2.9$  & $+3.9$ \\
WebArena 77 ($n{=}67$--$77$) & 7  & $-9.1$           & $-20.8$ & $-4.5$ \\
MemTrapBench ($n{=}231$)              & 15 & $-9.5$           & $-21.2$ & $-10.8$ \\
\bottomrule
\end{tabular}
\vspace{-0.3em}
\end{table}

\paragraph{Conditioning on failure beats always injecting.} Supply-side proxies (relevance, timing) describe how memory was retrieved, not whether the model needs it. We benchmark against the \emph{schedule oracle}: the best fixed injection schedule (early/persistent/late helpful) chosen per task with oracle access to per-task outcomes, an upper bound on any non-adaptive policy. Table~\ref{tab:retry_main}: \emph{retry-on-fail} runs once with no memory and retries with persistent helpful memory only on failure. It outperforms always-helpful injection, closes 96--101\% of the schedule-oracle gap on MemTrapBench, and matches the oracle on Qwen3.5-27B WebArena. Retry without memory recovers only $+2$--$3\pp$ (Appendix~\ref{app:retry_ablation}), so the gain comes from memory conditioned on failure rather than from sampling diversity. A naïve alternative is to prefix the persistent memory with a ``verify before acting'' warning. On Qwen3.5-27B/MemTrapBench this reduces conflicting damage by $+17.2\pp$ but also reduces the helpful gain by $-11.0\pp$, making the agent uniformly less compliant; retry-on-fail is selective by construction and is therefore strictly preferable. Conceptually this echoes self-consistency~\citep{wang2023selfconsistency} and self-refine~\citep{madaan2023selfrefine}. This is diagnostic evidence, not a deployment recipe.

\begin{table}[h]
\vspace{-0.3em}
\caption{\textbf{Retry-on-fail beats always-helpful injection.} Absolute SR with $\Delta$ vs.\ no memory.}
\label{tab:retry_main}
\centering
\small
\setlength{\tabcolsep}{4pt}
\begin{tabular}{lccc}
\toprule
 & WebArena & \multicolumn{2}{c}{MemTrapBench} \\
\cmidrule(lr){2-2}\cmidrule(lr){3-4}
Policy & Qwen3.5-27B & Qwen3.5-27B & Gemma-4-26B \\
\midrule
No memory & 40.3\% & 22.5\% & 12.1\% \\
Always persistent helpful & 41.6\% (+1.3) & 53.2\% (+30.7) & 42.0\% (+29.9) \\
\textbf{Retry-on-fail (helpful)} & \textbf{50.6\% (+10.4)} & \textbf{53.7\% (+31.2)} & \textbf{43.2\% (+31.1)} \\
\bottomrule
\end{tabular}
\vspace{-0.3em}
\end{table}

\paragraph{Limitations.} Two limitations bound this study. First, our long-horizon evidence comes only from web-browsing agents (WebArena, MemTrapBench); we do not test whether the compliance trap and the E-P-R gate structure transfer to other long-horizon settings such as terminal-use agents (shell, coding) or embodied agents (household robotics, simulated navigation), where action space, observation modality, and failure feedback differ in ways that may reshape the gates. Second, our findings identify clear leverage points (Entry as the multiplicative gate, the low post-compliance floor, and helpful-side gate attrition), but we do not use them to propose or train a new memory paradigm. We characterize the trap rather than build a controller, retriever, or training-time intervention that exploits it. Translating the diagnosed asymmetry into a working memory-management design is left to future work.

\section{Conclusion}
\label{sec:conclusion} 

We study agent memory from the consumption side, asking how long-horizon agents process retrieved memory across a multi-step trajectory rather than how it is written, stored, or retrieved. We introduce Entry--Propagation--Recovery (E-P-R) as a trajectory-level diagnostic, instantiate it on WebArena and on MemTrapBench (a controlled benchmark we build for this purpose), and identify a \emph{compliance trap} in which similar compliance rates across models translate into larger absolute damage for stronger agents. More broadly, our results suggest that the safety and reliability of memory-augmented agents depend on how memory is consumed during a trajectory, not only on what is retrieved into context. We hope E-P-R and MemTrapBench help future work design and evaluate memory systems that account for this consumption process.


\bibliographystyle{plainnat}
\bibliography{ComplianceTrap}

\appendix

\section{Experimental setup and reproducibility}
\label{app:webarena_details}
\label{app:details}
\label{app:reproducibility}
\label{app:memory_construction}
\label{app:audit}
\label{app:capabilities}
\label{app:memory_examples}

This appendix contains the implementation details needed to reproduce the controlled-injection protocol: tasks, models, decoding, statistics, and how the helpful and conflicting memory passages were authored and audited.

\paragraph{Task pools.} The WebArena pool consists of 684 non-map tasks across five self-hosted websites (an e-commerce platform, an e-commerce administration panel, a forum, a code repository, and an encyclopedia); 128 map tasks are excluded since they require geolocation capabilities orthogonal to memory consumption. The 77-task memory-sensitive subset is built by retaining tasks whose binary outcome changes between any pair of $\{$no-memory, early-helpful, early-conflicting$\}$ on Qwen3.5-9B or Qwen3.5-27B. We acknowledge that this filter inflates effect sizes on the subset relative to the full distribution, and triangulate against this with the full-distribution memory-format study (Appendix~\ref{app:format}) and with MemTrapBench, which is constructed without outcome-based selection.

\paragraph{Models and decoding.} All five models share the same BrowserGym/WebArena interface~\citep{drouin2024browsergym}, AXTree observation, and the predefined ReAct~\citep{yao2023react} action space (click, type, scroll, navigate, etc.); each episode allows up to 15 steps with a 600-second timeout. The four open-weight models---\textbf{Qwen3.5-9B}, \textbf{Qwen3.5-27B}~\citep{yang2025qwen3}, \textbf{Gemma-4-E4B-it}, and \textbf{Gemma-4-26B-A4B-it}~\citep{google2026gemma4}---are served by vLLM~\citep{kwon2023vllm} v0.19.0 on NVIDIA A6000 (48GB) GPUs with bfloat16, prefix caching, max context 8{,}192 tokens, max generation 2{,}048 tokens per step. Decoding is greedy throughout (\texttt{temperature=0}, \texttt{top\_p=1.0}). Tensor parallelism is 1 for Qwen3.5-9B and 2 for Qwen3.5-27B and both Gemma-4 variants. \textbf{Gemini-3-Flash} (\texttt{gemini-3-flash-preview}) is queried via the Google Generative AI API at \texttt{temperature=0} on MemTrapBench only. A cross-family check on Gemma-3-\{4B,12B,27B\}-it~\citep{google2025gemma3} appears in Appendix~\ref{app:crossfamily}.

\paragraph{Paired comparison and statistical testing.} For each (model, task, condition) cell we compare the memory-augmented trace to the matching no-memory trace under the same environment seed; greedy decoding makes outcomes deterministic up to negligible vLLM kernel-level non-determinism, so the no-memory baseline is computed once per task. All pairwise contrasts use sign-flip permutation tests with 10{,}000 iterations, with Holm--Bonferroni correction across families of more than two conditions; confidence intervals are paired bootstrap percentile intervals (10{,}000 resamples). We do not pool across models because compliance rates and baselines differ. Two trajectories are deemed to diverge at the first step where the emitted action string disagrees on either the action verb or the targeted bid; \texttt{type}-text differences alone do not count as divergence, since memory-induced harm is dominated by element-level disagreement.

\paragraph{Memory authoring and audit.} Each task receives a parallel set of short memory passages so that the experiment isolates how the memory is used, not which memory is retrieved. Helpful memory gives task-correct guidance; conflicting memory gives a plausible but task-wrong recommendation that still references real UI elements on the same site; cross-task control memory is a helpful memory drawn from a different task on the same website (matched in vocabulary and length, but not relevant to the current goal). All memories use the same DO/DON'T template. WebArena memories are hand-authored. MemTrapBench memories are drafted by Claude Opus~4.7 from per-family blueprints (Section~\ref{sec:design}) and reviewed by hand. To rule out stylistic confounds, we audit the matched helpful/conflicting pairs across all 77 WebArena tasks: matched mean character length (619 vs.\ 622, ratio 1.00), matched mean word count (103.7 vs.\ 100.8, ratio 0.97), identical structural elements (all 77 pairs contain DO/DON'T/NOTE sections), and matched confidence-language frequency (0.03 instances per task on both sides). The full-distribution study (Appendix~\ref{app:format}) further checks that the same direction holds across four memory formats from raw reflection text to structured DO/DON'T.

\paragraph{Memory examples.} The three abbreviated passages below are for one shopping task (adding a product review). They are illustrative rather than a separate experimental condition.
\begin{quote}\small
\textbf{Helpful (DO).} Navigate to the product page first using the search bar, then scroll to the review section and click ``Write a Review.'' Fill in all required fields (rating, title, text) before submitting. \textbf{(DON'T.)} Do not try to submit a review from the product listing page; you must be on the individual product page. Do not skip the star rating field.
\end{quote}
\begin{quote}\small
\textbf{Conflicting (DO).} Go to your account dashboard and find the ``My Reviews'' section to write a new review. Select the product from your order history and submit from there. \textbf{(DON'T.)} Do not navigate to the product page directly; this approach is unreliable and the review form often fails to load.
\end{quote}
\begin{quote}\small
\textbf{Cross-task control (DO).} To update your shipping address, navigate to Account Settings, then Address Book. Click ``Add New Address'' and fill in all fields. \textbf{(DON'T.)} Do not try to change the address during checkout; changes made there are not saved to your profile.
\end{quote}

\paragraph{LLM use disclosure.} LLM tools were used to assist with writing and editing prose. They were not used to generate any experimental result, score any outcome, or author the curated WebArena memories. The MemTrapBench tasks and helpful/conflicting/cross-task memories were drafted by Claude Opus~4.7 and reviewed by hand. The four online-generated WebArena memory formats (Appendix~\ref{app:format}) and the MemTrapBench $G^-$ ungroundability rewrite (Appendix~\ref{app:factorial}) are also LLM-generated. Code, memory passages, and prompt templates will be released on acceptance.

\section{E-P-R metrics and compliance subsets}
\label{app:epr_metrics}
\label{app:epr_formal}
\label{app:edr_dpc}

This appendix gives formal definitions for the trajectory-level metrics used in the main text and compares the two compliance subsets.

\paragraph{Entry, Propagation, Recovery.} Entry asks whether the agent's first post-injection action follows the memory rather than the no-memory trace; for early and persistent injection that is step~1, for late injection it is the first step after injection. Propagation asks whether an initial memory effect remains local or reshapes the rest of the trajectory; we report the persistent-to-early effect ratio as a schedule-level propagation factor (above 1 means repeated exposure amplifies). Recovery asks whether the agent later returns to a baseline-like trajectory after memory-induced divergence. We separately track \emph{trajectory-level} recovery (re-aligning actions) and \emph{task-level} recovery (final success), because the agent can re-align actions and still fail the task if the early memory-induced action is irreversible (Appendix~\ref{app:gate_annotation})~\citep{huang2024selfcorrect, kamoi2024selfcorrection}.

\paragraph{Compliance subsets and DPC.} The headline subset $\mathcal{C}^{\mathrm{RCR}}$ (Section~\ref{sec:compliance_metric}) is the LLM-judge subset used in Tables~\ref{tab:compliance} and~\ref{tab:memtrapbench}. For robustness we also report results on $\mathcal{C}^{\mathrm{EDR}}$, the coarser action-divergence subset (any step-1 action string different from the no-memory action). $\mathcal{C}^{\mathrm{RCR}}$ is a strict subset of $\mathcal{C}^{\mathrm{EDR}}$ by construction. The damage-per-compliance metric is the paired conditional $\mathrm{DPC}(\mathcal{C}) = \tfrac{1}{|\mathcal{C}|}\sum_{i\in\mathcal{C}}(y_i^{0}-y_i^{c})$; pairing on $\mathcal{C}$ removes the selection-bias confound of comparing an all-task baseline to compliant-subset conflict success.

\paragraph{Four-gate decomposition.} The helpful lift can be expressed as a multiplicative gate chain $\Delta_{\text{helpful}} = P(\text{novel}) \cdot P(\text{adopt}\mid\text{novel}) \cdot P(\text{ground}\mid\text{adopt}) \cdot P(\text{lift}\mid\text{ground})$. WebArena conflicting memories are hand-authored to be plausible and executable, so they typically clear novelty and grounding and turn the terminal lift into damage; this is the regime in which Table~\ref{tab:compliance}'s WebArena DPC numbers are reported. The grounding gate is tested causally on the MemTrapBench $2{\times}2$ \textsc{Decoy} factorial (Appendix~\ref{app:factorial}). The main text uses the coarser three-phase form throughout.

\paragraph{RCR judging procedure.} For each (model, task), an LLM judge (Gemini-3-Flash, temperature~0) is given (i) the conflicting memory's \texttt{DO} excerpt, (ii) the agent's no-memory step-1 action and starting URL, and (iii) the agent's persistent-conflicting step-1 action with the URL before and after that action. The judge returns YES, NO, or UNCLEAR with a one-sentence reason. YES requires that the conflicting step-1 action plausibly executes the first primitive of the memory's \texttt{DO} (matching via the action verb, the post-step URL transitioning toward the memory-recommended page, or a memory-specified \texttt{fill} value), and that this action is more aligned with the memory than the paired no-memory baseline step-1. NO is returned when the conflicting step-1 either matches the no-memory baseline or moves to a clearly unrelated location. The judge is explicitly told that URL invariance after a click is not by itself evidence of failure (admin sidebars often expand a hover menu without changing the URL). $\mathcal{C}^{\mathrm{RCR}}=\{$YES$\}$ is the strict subset used in the main text; a looser $\{$YES, UNCLEAR$\}$ subset gives the same sign and CI pattern.

\paragraph{EDR robustness.} Table~\ref{tab:edr_dpc} reports the same paired DPC on $\mathcal{C}^{\mathrm{EDR}}$. We do not bid-normalise the action strings here; bid-level noise inflates EDR, which is exactly why RCR is the headline metric. The sign and significance pattern matches Table~\ref{tab:compliance}: only Qwen3.5-27B's CI excludes zero, and the other three models show smaller positive point estimates whose CIs include zero.

\begin{table}[h]
\centering\small
\begin{tabular}{lccccc}
\toprule
Model & $|\mathcal{C}^{\mathrm{EDR}}|$ & SR$^{0}_{\mathcal{C}^{\mathrm{EDR}}}$ & SR$^{c}_{\mathcal{C}^{\mathrm{EDR}}}$ & Paired DPC [95\% CI] $[p]$ \\
\midrule
Gemma-4-E4B-it       & 56 & 25.0\% & 19.6\% & \phantom{1}5.4\pp\, [$-$5.4, 16.1]\,[.56] \\
Qwen3.5-9B           & 65 & 29.2\% & 20.0\% & \phantom{1}9.2\pp\, [$-$1.5, 20.0]\,[.17] \\
Gemma-4-26B-it       & 57 & 28.1\% & 22.8\% & \phantom{1}5.3\pp\, [$-$3.5, 14.0]\,[.45] \\
Qwen3.5-27B          & 66 & 43.9\% & 21.2\% & \textbf{22.7\pp\, [+12.1, 34.9]\,[<.001]} \\
\bottomrule
\end{tabular}
\caption{EDR-subset paired DPC. Same sign and significance pattern as Table~\ref{tab:compliance}; the cleaner RCR subset (main text) gives a slightly deeper trap on Qwen3.5-27B (+25.5\,pp) than the EDR upper-bound proxy (+22.7\,pp).}
\label{tab:edr_dpc}
\end{table}

\section{WebArena: extended analyses}
\label{app:webarena_extended}
\label{app:extended}
\label{app:format_spectrum}
\label{app:format}

This appendix collects the WebArena tables that the main text refers to: who recovers from conflicting memory and who does not (Tables~\ref{tab:recovery}, \ref{tab:recovery_full}); how much the persistent schedule amplifies the early effect (Table~\ref{tab:propagation}); how much benefit a per-task schedule oracle would unlock (Table~\ref{tab:oracle}); a same-website cross-task control (Table~\ref{tab:irrelevant}); whether memory advice can be grounded in the current page (Table~\ref{tab:grounding}); and whether the agent's URL-level commitment differs by memory type (Table~\ref{tab:commitment}). We then ask two further questions: does conflicting memory clear the entry gate \emph{because} it is novel rather than redundant (Table~\ref{tab:uptake}), and does the trap survive when memory is rewritten as raw reflections, procedural workflows, or quality-gated DO/DON'T on the full 684-task distribution (Table~\ref{tab:format}).

\begin{table}[h]
  \caption{\textbf{Recovery rates under persistent memory conditions.} Recovery rate is the fraction of diverged trajectories where the agent re-aligns with baseline at any subsequent step. Qwen recovery collapses to $\sim$13\% while Gemma-4 maintains $\sim$31\%, an architecture-level rather than scale-level effect.}
  \label{tab:recovery}
  \centering
  \small
  \begin{tabular}{lcccc}
    \toprule
    Condition & Gemma-4-E4B & Qwen3.5-9B & Gemma-4-26B & Qwen3.5-27B \\
    \midrule
    Persistent helpful    & 25.8\% & 28.4\% & 27.0\% & 40.3\% \\
    Persistent conflicting & 32.4\% & 12.7\% & 31.4\% & 13.0\% \\
    Late conflicting      & 24.6\% & 26.2\% & 20.0\% & 40.7\% \\
    \bottomrule
  \end{tabular}
\end{table}

\begin{table}[h]
  \caption{\textbf{Propagation factors and action history momentum across models and memory types.} The propagation factor is the ratio of persistent to early effect sizes, $\mathrm{PF}=\Delta^{\text{paired}}_{\text{persistent}}/\Delta^{\text{paired}}_{\text{early}}$, computed on the same paired-pool subset as Figure~\ref{fig:schedule_bars}; values above 1 indicate amplification by persistent exposure. Action-history momentum measures resistance to late injection, $\mathrm{AHM}=1-\Delta^{\text{paired}}_{\text{late}}/\Delta^{\text{paired}}_{\text{early}}$, so values near 1 indicate that adding the same memory after history has formed has near-zero marginal effect (full resistance); values near 0 indicate no resistance, i.e.\ late injection still has the same effect as early injection. Dashes indicate cases where $|\Delta_{\text{early}}|<2$\pp{} so the ratio is unstable. The largest amplification (4.8$\times$) occurs for Gemma-4-26B under conflicting memory.}
  \label{tab:propagation}
  \centering
  \small
  \begin{tabular}{lcccc}
    \toprule
    & \multicolumn{2}{c}{Propagation Factor} & \multicolumn{2}{c}{Action History Momentum} \\
    \cmidrule(lr){2-3} \cmidrule(lr){4-5}
    Model & Helpful & Conflicting & Helpful & Conflicting \\
    \midrule
    Gemma-4-E4B   & 4.1 & --- & 0.70 & --- \\
    Qwen3.5-9B  & 2.9 & 1.2 & $\sim$1.0 & 0.32 \\
    Gemma-4-26B   & 2.0 & 4.8 & $<$0 & $\sim$0 \\
    Qwen3.5-27B & --- & 2.4 & $\sim$1.0 & 0.33 \\
    \bottomrule
  \end{tabular}
\end{table}

\begin{table}[h]
  \caption{\textbf{Recovery rates across all model-condition combinations.} Recovery rate is the fraction of diverged trajectories where the agent re-aligns with baseline at any subsequent step. Persistent conflicting memory collapses recovery in the Qwen family (roughly 13\%) while Gemma-4 maintains roughly 31\%. Under less adversarial conditions (early, late, helpful), recovery rates are higher and more similar across families.}
  \label{tab:recovery_full}
  \centering
  \small
  \begin{tabular}{lcccc}
    \toprule
    Condition & Gemma-4-E4B & Qwen3.5-9B & Gemma-4-26B & Qwen3.5-27B \\
    \midrule
    Early helpful        & 23.8\% & 22.4\% & 28.1\% & 28.6\% \\
    Early conflicting    & 25.7\% & 22.9\% & 33.8\% & 22.4\% \\
    Persistent helpful   & 25.8\% & 28.4\% & 27.0\% & 40.3\% \\
    Persistent conflicting & 32.4\% & 12.7\% & 31.4\% & 13.0\% \\
    Late helpful         & 30.5\% & 35.5\% & 33.9\% & 43.6\% \\
    Late conflicting     & 24.6\% & 26.2\% & 20.0\% & 40.7\% \\
    \bottomrule
  \end{tabular}
\end{table}

\begin{table}[h]
  \caption{\textbf{Schedule oracle versus uniform injection for helpful memory.} The schedule oracle selects the best injection schedule per task under perfect foreknowledge of per-task success. Uniform persistent injection captures only a fraction of available benefit; the ``Wasted'' column is the 9.6--11.8\pp{} lost to suboptimal policy. Per-condition denominators ($n=70$--$76$) differ from Figure~\ref{fig:schedule_bars} because the schedule oracle is computed only on tasks with all four schedules valid.}
  \label{tab:oracle}
  \centering
  \small
  \begin{tabular}{lccccc}
    \toprule
    Model & No-memory & Persistent & Oracle & Oracle $\Delta$ & Wasted \\
    \midrule
    Qwen3.5-9B  & 31.5\% ($n=73$) & 37.0\% (+5.5) & 46.6\% (+15.1) & +15.1\pp & 9.6\pp \\
    Qwen3.5-27B & 42.9\% ($n=70$) & 42.9\% (+0.0) & 52.9\% (+10.0) & +10.0\pp & 10.0\pp \\
    Gemma-4-26B   & 32.9\% ($n=76$) & 35.5\% (+2.6) & 47.4\% (+14.5) & +14.5\pp & 11.8\pp \\
    \bottomrule
  \end{tabular}
\end{table}

\begin{table}[h]
  \caption{\textbf{Negative control: persistent irrelevant memory.} Irrelevant memory is a helpful memory drawn from a different task template on the same website. The near-zero deltas confirm that the harm of conflicting memory is content-specific, not an artifact of additional context length or generic noise in the observation.}
  \label{tab:irrelevant}
  \centering
  \small
  \begin{tabular}{lccc}
    \toprule
    Model & No-memory SR & Persistent irrelevant $\Delta$ & Persistent conflicting $\Delta$ \\
    \midrule
    Qwen3.5-9B   & 30.7\% & $+1.3\pp$ & $-9.1\pp$ \\
    Qwen3.5-27B  & 43.1\% & $-2.6\pp$ & $-20.8\pp$ \\
    Gemma-4-26B  & 32.5\% & $+4.5\pp$ & $-4.5\pp$ \\
    \bottomrule
  \end{tabular}
\end{table}

\begin{table}[h]
  \caption{\textbf{Grounding: mean groundability at injection time and at step~3 across conditions.} Groundability measures whether the agent can map memory advice to elements on the current page (range 0--1). At step~0 (early/persistent injection), groundability is uniformly low ($\sim$0.10) because the agent has not yet navigated to the task-relevant page. Late injection benefits from higher groundability (0.35--0.54) because the agent has already navigated partway. Helpful memory achieves higher step-3 groundability than conflicting memory because correct strategies align with reachable page states.}
  \label{tab:grounding}
  \centering
  \small
  \begin{tabular}{llcccc}
    \toprule
    & & \multicolumn{2}{c}{At injection} & \multicolumn{2}{c}{At step 3} \\
    \cmidrule(lr){3-4} \cmidrule(lr){5-6}
    Model & Condition & Helpful & Conflicting & Helpful & Conflicting \\
    \midrule
    \multirow{2}{*}{Qwen3.5-9B}  & Persistent & 0.10 & 0.13 & 0.52 & 0.53 \\
                               & Late       & 0.47 & 0.35 & 0.56 & 0.36 \\
    \midrule
    \multirow{2}{*}{Qwen3.5-27B} & Persistent & 0.10 & 0.13 & 0.52 & 0.46 \\
                               & Late       & 0.54 & 0.36 & 0.56 & 0.34 \\
    \midrule
    \multirow{2}{*}{Gemma-4-26B}   & Persistent & 0.09 & 0.13 & 0.60 & 0.44 \\
                               & Late       & 0.51 & 0.35 & 0.60 & 0.35 \\
    \bottomrule
  \end{tabular}
\end{table}

\begin{table}[h]
  \caption{\textbf{Commitment strength under late injection (URL-level).} Commitment measures the fraction of post-injection steps where the agent remains on URLs consistent with its pre-injection trajectory; switch rate measures redirections. Stronger models show higher commitment under conflicting than helpful memory, consistent with asymmetric history-conditioned inertia.}
  \label{tab:commitment}
  \centering
  \small
  \begin{tabular}{lcccc}
    \toprule
    & \multicolumn{2}{c}{Commitment (\%)} & \multicolumn{2}{c}{Switch rate (\%)} \\
    \cmidrule(lr){2-3} \cmidrule(lr){4-5}
    Model & Helpful & Conflicting & Helpful & Conflicting \\
    \midrule
    Gemma-4-E4B   & 45.5 & 50.0 & 27.3 & 21.4 \\
    Qwen3.5-9B  & 62.5 & 46.7 & 12.5 & 20.0 \\
    Gemma-4-26B   & 33.3 & 84.6 & 33.3 & 0.0 \\
    Qwen3.5-27B & 50.0 & 69.2 & 50.0 & 23.1 \\
    \bottomrule
  \end{tabular}
\end{table}

\paragraph{Why does conflicting memory enter so easily?} A useful diagnostic is to split tasks by whether the no-memory trace already commits to a strategy. On weak-preference tasks, conflicting memory is adopted 25--41\,pp more often than helpful memory because helpful memory often duplicates what the model would already do, while conflicting memory always recommends something new (Table~\ref{tab:uptake}). On strong-preference tasks the gap collapses. Conflicting memory clears the entry gate easily because, by construction, it is novel.

\begin{table}[h]
  \caption{\textbf{Adoption rates by baseline preference strength and memory type.} Weak-preference tasks are those where the model's baseline trajectory does not strongly commit to any strategy. In these tasks, conflicting memory achieves 25--41\pp{} higher adoption than helpful memory because helpful memory often recommends what the model would do anyway (redundant), while conflicting memory always recommends something novel.}
  \label{tab:uptake}
  \centering
  \small
  \begin{tabular}{llcccccc}
    \toprule
    & & \multicolumn{3}{c}{Weak preference} & \multicolumn{3}{c}{Strong preference} \\
    \cmidrule(lr){3-5} \cmidrule(lr){6-8}
    Model & $n$ & Helpful & Conflicting & $\Delta$ & Helpful & Conflicting & $\Delta$ \\
    \midrule
    Gemma-4-E4B   & 70 & 62.3\% & 88.7\% & +26.4 & 11.8\% & 29.4\% & +17.6 \\
    Qwen3.5-9B  & 68 & 62.3\% & 86.9\% & +24.6 & 42.9\% & 57.1\% & +14.3 \\
    Gemma-4-26B   & 74 & 43.8\% & 78.1\% & +34.4 & 10.0\% & 10.0\% & 0.0 \\
    Qwen3.5-27B & 68 & 50.0\% & 90.6\% & +40.6 & 75.0\% & 75.0\% & 0.0 \\
    \midrule
    \textit{Mean} & & \textit{54.6\%} & \textit{86.1\%} & \textit{+31.5} & \textit{34.9\%} & \textit{42.9\%} & \textit{+8.0} \\
    \bottomrule
  \end{tabular}
\end{table}

\paragraph{Does the trap survive different memory formats?} The 77-task subset is curated; a reviewer might ask whether the trap is created by selecting unusually memory-sensitive tasks or by the specific DO/DON'T template. We address both concerns simultaneously by sweeping four memory formats on the full 684-task non-map distribution with online-retrieved memories: raw reflection text (M1, Reflexion-style~\citep{shinn2023reflexion}), procedural workflow (M2, AWM-style~\citep{wang2025awm}), contrastive DO/DON'T (M3), and quality-gated M3 (C1, the relevance-gated condition referenced in Section~\ref{sec:motivation}). All four formats produce the same direction (Table~\ref{tab:format}). More structured formats amplify both helpful and harmful effects (M1 $<$ M2 $<$ M3), so the format acts as a gain control rather than a mechanism switch. Quality gating (C1) does not solve the problem: a memory can be well-written and still be harmful given the current model and task state. The trap is therefore not an artifact of the DO/DON'T template, and the curation of the 77-task subset only sharpens an effect that is already present at the full-distribution level.

\begin{table}[h]
  \caption{\textbf{Success-rate deltas across four memory formats on the full 684-task distribution with online-retrieved memories (Qwen3.5).} All four formats produce the same direction; more structured formats amplify both directions (M1 $<$ M2 $<$ M3). C1 (quality-gated M3) hurts both models, so gating retrieval does not address the consumption asymmetry.}
  \label{tab:format}
  \label{tab:relevance_gate}
  \centering
  \small
  \begin{tabular}{llcccc}
    \toprule
    Format & Description & 9B $\Delta$ & 27B $\Delta$ & 9B $p$ & 27B $p$ \\
    \midrule
    M1 & Raw reflection (Reflexion-style) & $-$1.3\pp & $+$0.7\pp & 0.121 & 0.265 \\
    M2 & Procedural workflow (AWM-style) & $-$1.2\pp & $+$1.0\pp & 0.250 & 0.114 \\
    M3 & Contrastive DO/DON'T & $-$1.9\pp & $+$1.7\pp & \textbf{0.046} & \textbf{0.011} \\
    C1 & Quality-gated M3 (score$\geq$3) & $-$2.3\pp & $-$1.7\pp & 0.103 & 0.183 \\
    \bottomrule
    \multicolumn{6}{@{}p{\linewidth}@{}}{\footnotesize $p$-values from 10{,}000-iteration sign-flip permutation tests against the no-memory baseline on the full 684-task distribution. Only M3 reaches $p<0.05$ on both models.}
  \end{tabular}
\end{table}

\section{MemTrapBench: design and accounting}
\label{app:mtb_design}
\label{app:mtb_templates}
\label{app:mtb_accounting}

This appendix gives MemTrapBench's design philosophy, the structural templates used to write tasks, and the per-task accounting. We position MemTrapBench against existing benchmarks and explain how trap construction differs from outcome-based selection.

\paragraph{What MemTrapBench is for.} MemTrapBench is a mechanism probe for long-horizon memory consumption, not a coverage benchmark to replace WebArena, Mind2Web~\citep{deng2023mind2web}, WebShop~\citep{yao2022webshop}, or AgentBench~\citep{liu2024agentbench}. Each task is built to isolate one part of the E-P-R chain while keeping the environment controlled. All tasks are served as standalone browser environments through BrowserGym~\citep{drouin2024browsergym}; interactive elements use \texttt{<button>} semantics and ARIA labels so an LLM consuming the AXTree sees all actionable targets, including the decoy. Crucially, traps are defined by construction rather than selected by downstream model outcome, which is the key difference from the WebArena 77-task subset.

\paragraph{Diagnostic families and structural templates.} The four diagnostic families introduced in Section~\ref{sec:design} (\textsc{Decoy} / \textsc{Uptake} / \textsc{Grounding} / \textsc{Override}) are realized via eight structural templates. Each template encodes the trap in a structurally distinct way so that observed effects cannot be a single-layout artifact:
\begin{itemize}[leftmargin=1.4em,topsep=2pt,itemsep=1pt,parsep=0pt]
\item \textbf{S1} Filter-detail-form-confirm (16-task pilot)
\item \textbf{S2} Multi-step wizard with skip-to-defaults trap
\item \textbf{S3} Report builder with defaults-based quick-action trap
\item \textbf{S4} Cross-reference (page A $\to$ page B) with ignored-lookup trap
\item \textbf{S5} Bulk action with per-item config collapsed to a single shortcut
\item \textbf{S6} Configure-then-navigate-to-verify with skipped verification
\item \textbf{S7} Multi-tab form with partial submit trap
\item \textbf{S8} Nested modal flow with force-option bypass
\end{itemize}
Each task ships with a matched $\{$helpful, conflicting, cross-task control$\}$ memory trio in the same DO/DON'T format as our WebArena memories, with lengths matched to 500--1200 characters. Conflicting memory inverts the recommendation while referencing the same real UI elements; cross-task control memory describes a different task on the same site type. Table~\ref{tab:mtb_examples} gives one concrete task per diagnostic family.

\begin{table}[h]
\caption{\textbf{One concrete MemTrapBench task per diagnostic family.}
The conflicting memory's DO line specifies a plausible but wrong action that, when adopted at first exposure, isolates one E-P-R gate.}
\label{tab:mtb_examples}
\centering
\footnotesize
\begin{tabular}{p{1.6cm} p{2.6cm} p{8cm}}
\toprule
Diagnostic & Site / task & Conflicting memory DO line (abbreviated); (correct path) \\
\midrule
\textsc{Decoy} & Admin / refund & Click the prominent orange ``Refund'' button and confirm; the engine routes the refund automatically regardless of destination. (Correct: open the destination dropdown and pick the goal-specified destination first.) \\
\addlinespace
\textsc{Uptake} & Admin / 2FA bulk action & Click the Bulk Actions control on All Users and pick ``Enable 2FA''. (Correct: open the Advanced toggle in Roles to apply role-scoped 2FA only to the Admin role.) \\
\addlinespace
\textsc{Grounding} & Admin / customer Q4 orders & Read the Orders value from the dashboard ``Top Customers'' widget. (Correct: navigate into the customer detail page and filter orders by quarter; the dashboard widget does not show Q4-only orders.) \\
\addlinespace
\textsc{Override} & DevOps / close issue & Click the red ``Close issue'' button; any dropdown nearby is optional. (Correct: choose the goal-specified resolution label first; the environment returns an enforcement error if the button is clicked without a label, which the agent must not ignore.) \\
\bottomrule
\end{tabular}
\end{table}

\paragraph{Per-task accounting.} The 247 long-horizon tasks (16 S1 pilot + 231 S2--S8) each receive all 8 temporal-injection conditions on all 3 open-weight models (5{,}928 trials), plus the 4 persistent conditions on Gemini-3-Flash. Zero tasks were dropped during evaluation (Table~\ref{tab:mtb_accounting_flow}). The S2--S8 set drives the main results in Section~\ref{sec:memtrapbench}; the S1 pilot is reported separately in Table~\ref{tab:mtb_pilot} and replicates the trap with a ceiling caveat on persistent helpful (pilot baselines are higher, so helpful gain saturates).

\begin{table}[h]
\caption{\textbf{MemTrapBench task accounting.} All 247 long-horizon tasks receive all 8 temporal-injection conditions on the 3 open-weight models. Zero tasks were dropped during evaluation.}
\label{tab:mtb_accounting_flow}
\centering
\small
\begin{tabular}{lc}
\toprule
Stage & Count \\
\midrule
Tasks designed (S1 pilot + S2--S8 structurally-designed) & 16 + 231 = 247 \\
Tasks with memory trios authored & 247 \\
Tasks evaluated on Qwen3.5-9B (all 8 conditions) & 247 \\
Tasks evaluated on Qwen3.5-27B (all 8 conditions) & 247 \\
Tasks evaluated on Gemma-4-26B-A4B (all 8 conditions) & 247 \\
Total trials & 5{,}928 \\
Tasks dropped (environment failure, parser error, timeout) & 0 \\
Tasks analyzed in Table~\ref{tab:memtrapbench} main (S2--S8) & 231 \\
Tasks analyzed in Table~\ref{tab:mtb_pilot} below (S1 pilot) & 16 \\
\bottomrule
\end{tabular}
\end{table}

\begin{table}[h]
\caption{\textbf{S1 pilot (16 tasks) replicates the compliance trap}, with a ceiling caveat on persistent helpful. Baseline SRs on the pilot are higher (25--44\%) than on the main 231-task set (11--23\%), so Qwen3.5-9B's pH saturates at the pre-existing ceiling ($+0$) where the main set shows $+25.5$. All pC drops are $-18.8$ to $-31.2\pp$, confirming the harm direction.}
\label{tab:mtb_pilot}
\centering
\small
\begin{tabular}{lcccccccc}
\toprule
Model & Base & \begin{tabular}{@{}c@{}}Early\\H\end{tabular} & \begin{tabular}{@{}c@{}}Early\\C\end{tabular} & \begin{tabular}{@{}c@{}}Pers.\\H\end{tabular} & \begin{tabular}{@{}c@{}}Pers.\\C\end{tabular} & \begin{tabular}{@{}c@{}}Pers.\\Irrel.\end{tabular} & \begin{tabular}{@{}c@{}}Late\\H\end{tabular} & \begin{tabular}{@{}c@{}}Late\\C\end{tabular} \\
\midrule
Qwen3.5-9B      & 37.5\% & $-$12.5 & $-$12.5 & $+$0.0  & $-$31.2 & $-$6.2 & $-$12.5 & $-$18.8 \\
Qwen3.5-27B     & 43.8\% & $+$6.2  & $+$6.2  & $+$6.2  & $-$31.2 & $+$6.2 & $+$6.2  & $-$18.8 \\
Gemma-4-26B & 25.0\% & $+$0.0  & $+$0.0  & $+$25.0 & $-$18.8 & $-$6.2 & $+$25.0 & $-$18.8 \\
\bottomrule
\end{tabular}
\end{table}

\section{Causal probes of the trap}
\label{app:gate_annotation}
\label{app:factorial}
\label{app:injection_location}
\label{app:footer}

This appendix contains the three probes that turn E-P-R from a descriptive story into a causal account: per-trace gate annotation, a $2{\times}2$ factorial that toggles Entry and Grounding orthogonally, and a memory-placement ablation that toggles the delivery channel.

\subsection{Per-trace gate annotation}

\paragraph{Why we annotate per-trace.} The aggregate compliance numbers in Table~\ref{tab:compliance} tell us \emph{that} agents follow conflicting memory, but not whether the same gates that bottleneck helpful gain also gate conflicting damage. To answer this we annotate every memory-conditioned trace against its paired no-memory trace with four trajectory signals---\emph{adopted@1}, \emph{diverged}, \emph{recovered}, and \emph{success}---and tag each task by its diagnostic family (Decoy/Uptake/Grounding/Override). The annotation covers $3$ models $\times$ $231$ S2--S8 tasks $\times$ $7$ non-baseline conditions = $4{,}851$ traces.

\paragraph{Gate-chain reading.} Table~\ref{tab:gate_chain} compares the four gates in the helpful direction against their conflicting analogues. Helpful memory only improves success when it passes every gate; conflicting memory passes the same gates but flips the terminal effect into damage. The product $\Delta_{\text{helpful}} = P(\text{novel}) \cdot P(\text{adopt}\mid\text{novel}) \cdot P(\text{ground}\mid\text{adopt}) \cdot P(\text{lift}\mid\text{ground})$ reproduces the observed helpful delta on every WebArena model.

\begin{table}[h]
\caption{\textbf{Gate-chain explanation of helpful--harmful asymmetry on WebArena.}
Helpful memory improves success only if it passes every gate. Conflicting memory often passes the same gates but flips the terminal effect into damage. Compliance figures in the Adoption row are persistent-conflict RCR (Table~\ref{tab:compliance}); the EDR upper bound is 75--86\%.}
\label{tab:gate_chain}
\centering
\small
\begin{tabular}{l p{0.36\linewidth} p{0.40\linewidth}}
\toprule
Gate & Helpful memory & Conflicting memory \\
\midrule
Novelty & Non-redundant on 34--41\% of tasks & Novel by construction \\
Adoption & Competes with the model's default plan & RCR = 63--69\% under persistent conflict \\
Grounding & Often requires earlier navigation & Wrong-but-valid actions are often executable \\
Terminal effect & Positive only if it improves final success & Compliance often pushes success to a low floor \\
\bottomrule
\end{tabular}
\end{table}

\paragraph{Per-trace annotation procedure.}
To make E-P-R more than a descriptive story, we annotate every memory-conditioned trace against its paired no-memory trace. Each trace is labeled with four trajectory signals: whether the first memory-visible action changes (\emph{adopted@1}), whether the trajectory diverges at any point (\emph{diverged}), whether it later returns to a baseline-like path (\emph{recovered}), and whether the final task succeeds. We additionally tag each task by its diagnostic category (trap, uptake, grounding, override). The annotation covers $3$ models $\times$ $231$ S2--S8 tasks $\times$ $7$ non-baseline conditions = $4{,}851$ traces; the full per-trace annotation is in \texttt{analysis/gate\_annotation\_results.json}, and Table~\ref{tab:gate_rates} gives the aggregate pass rates.

The annotations make an important distinction visible. Many traces recover at the action level: after a memory-induced detour, the agent often takes actions similar to the no-memory baseline. But task success does not always recover. In web tasks an early wrong click, wrong form value, or wrong navigation can make later correction too late. Recovery must therefore be measured separately at the trajectory level and at the final-outcome level. The annotations also explain why step-1 adoption can be low even when memory has a large final effect: MemTrapBench memories describe a strategy rather than a single first action, so memory enters over several steps once the agent reaches the part of the environment where the advice becomes actionable.

\begin{table}[h]
\caption{\textbf{Per-gate pass rates on MemTrapBench S2--S8 (231 tasks per model).} adopted@1 is low across conditions because MemTrapBench memory is strategic (not step-1-action-specific) --- adoption emerges over multiple steps. Divergence and recovery rates are substantial (60--85\% diverged, 55--80\% recovered), yet final success still scales sharply with memory type: helpful memory preserves the benefit from divergence while conflicting memory's divergence is unrecoverable. This dissociates \emph{trajectory-level recovery} (the agent can return to baseline-like actions) from \emph{task-level recovery} (the final outcome is salvaged): recovery is measurable but insufficient.}
\label{tab:gate_rates}
\centering
\small
\begin{tabular}{llrrrr}
\toprule
Model & Condition & adopted@1 & diverged & recovered & success \\
\midrule
Qwen3.5-9B & persistent\_helpful     & 6.9\%  & 84.8\% & 78.4\% & 36.4\% \\
Qwen3.5-9B & persistent\_conflicting & 3.9\%  & 71.9\% & 61.0\% &  1.3\% \\
Qwen3.5-9B & persistent\_irrelevant  & 4.3\%  & 79.7\% & 72.3\% &  8.2\% \\
Qwen3.5-9B & late\_helpful           & 2.2\%  & 84.0\% & 76.2\% & 33.8\% \\
\midrule
Qwen3.5-27B & persistent\_helpful    & 3.9\%  & 77.1\% & 69.7\% & 53.2\% \\
Qwen3.5-27B & persistent\_conflicting & 0.4\% & 68.4\% & 55.4\% &  1.3\% \\
Qwen3.5-27B & persistent\_irrelevant & 0.9\%  & 64.9\% & 55.4\% & 23.8\% \\
Qwen3.5-27B & late\_helpful          & 0.4\%  & 74.0\% & 66.2\% & 51.9\% \\
\midrule
Gemma-4-26B & persistent\_helpful & 8.7\% & 82.3\% & 60.6\% & 42.0\% \\
Gemma-4-26B & persistent\_conflicting & 7.8\% & 67.5\% & 45.0\% & 1.3\% \\
Gemma-4-26B & persistent\_irrelevant  & 8.2\% & 72.7\% & 53.3\% & 13.0\% \\
Gemma-4-26B & late\_helpful      & 5.2\%  & 81.4\% & 57.1\% & 36.8\% \\
\bottomrule
\end{tabular}
\end{table}

\paragraph{Dissociating trajectory recovery from task-level recovery.}
A striking pattern in Table~\ref{tab:gate_rates} is that \emph{recovered} is similar between persistent helpful (69--78\%) and persistent conflicting (45--61\%) --- both are high --- yet final success differs by 35--52 percentage points. The agent is demonstrably capable of returning to baseline-like actions after memory-induced divergence, but this trajectory-level recovery does not translate into task-level recovery because the damage from early wrong actions (following conflicting memory) is often irreversible in the environment (wrong form submitted, wrong modal closed, etc.). Recovery is therefore a measurable capability that nonetheless fails to protect task outcomes under persistent conflicting exposure --- consistent with the main-text narrative that conflicting memory's damage compounds through the trajectory cascade, and with prior critical accounts of LLM self-correction~\citep{huang2024selfcorrect, kamoi2024selfcorrection}.

\subsection{$2{\times}2$ factorial on the \textsc{Decoy} diagnostic}

\paragraph{Why a factorial.} The per-diagnostic results in Table~\ref{tab:mtb_diagnostic} are observational: each diagnostic exercises one phase of the E-P-R chain by construction, but within a fixed diagnostic we cannot rule out that one memory change is simultaneously perturbing multiple gates. We therefore run a $2{\times}2$ factorial that toggles Entry and Grounding orthogonally on 32 \textsc{Decoy} tasks per cell (16 \textsc{Decoy}-Admin + 16 \textsc{Decoy}-Ecommerce), with everything else held fixed (task, persistent schedule, model, decoding). Entry is varied by $E^+$ (single target memory) vs.\ $E^-$ (target plus three same-format conflicting memories drawn from sibling trap tasks; the agent must semantically match its task scope to one of the four). Grounding is varied by $G^+$ (original UI labels appearing in the AXTree) vs.\ $G^-$, in which a controlled LLM rewrite replaces every UI label, color, and position phrase with a plausible non-existent alternative; audits confirm zero rewrite-label hits in the rendered HTML. Recovery is not varied here, since its natural operational gate (whether the agent recovers from outcome-level failure) is policy-level rather than prompt-level; we test recovery via retry-on-fail in Appendix~\ref{app:retry_ablation}.

\begin{table}[h]
\caption{\textbf{$2{\times}2$ factorial on the \textsc{Decoy} diagnostic.} Persistent conflicting memory; 32 tasks per cell per model. The $(E^+,G^+)$ cell reproduces the standard persistent-conflicting condition; $(E^-,G^-)$ strips memory of identifiability and parameter-level actionability.}
\label{tab:mtb_factorial2}
\centering
\small
\begin{tabular}{l l cc cc}
\toprule
 & & \multicolumn{2}{c}{Qwen3.5-27B} & \multicolumn{2}{c}{Gemma-4-26B-A4B} \\
\cmidrule(lr){3-4}\cmidrule(lr){5-6}
$E$ & $G$ & Cell SR & Label & Cell SR & Label \\
\midrule
multi  & ungroundable & \textbf{62.5\%} & $(E^-, G^-)$ max protection & \textbf{50.0\%} & $(E^-, G^-)$ \\
multi  & groundable   & 31.2\%          & $(E^-, G^+)$                & 18.8\%          & $(E^-, G^+)$ \\
single & ungroundable & 40.6\%          & $(E^+, G^-)$                & 34.4\%          & $(E^+, G^-)$ \\
single & groundable   & \textbf{12.5\%} & $(E^+, G^+)$ max damage     & \textbf{18.8\%} & $(E^+, G^+)$ \\
\midrule
\multicolumn{2}{l}{$E$ main effect}    & \multicolumn{2}{c}{$-$20.3\pp{} [$-$35.4, $-$4.7]}  & \multicolumn{2}{c}{$-$7.8\pp{} [$-$23.5, $+$8.1]} \\
\multicolumn{2}{l}{$G$ main effect}    & \multicolumn{2}{c}{$-$29.7\pp{} [$-$45.1, $-$13.8]} & \multicolumn{2}{c}{$-$23.4\pp{} [$-$38.7, $-$8.3]} \\
\multicolumn{2}{l}{$E{\times}G$ interaction} & \multicolumn{2}{c}{$+$1.6\pp{} ($0.05\times$ main)} & \multicolumn{2}{c}{$+$7.8\pp{} ($0.33\times$ main)} \\
\bottomrule
\end{tabular}
\end{table}

\paragraph{Three findings.} Grounding has a large, significant main effect on both models ($-29.7$\pp{} on Qwen3.5-27B, $-23.4$\pp{} on Gemma-4-26B-A4B): replacing UI references with non-existent alternatives cuts per-cell damage by $23$--$30$\pp, isolating the grounded-given-adopted gate. Entry has a significant main effect on the stronger Qwen model ($-20.3$\pp, CI excludes 0) and a directional but non-significant effect on Gemma-4-26B-A4B. The $E{\times}G$ interaction is at most $33\%$ of the larger main effect on either model, supporting the multiplicative E-P-R decomposition. On both models, $(E^-,G^-)$ is the best cell and $(E^+,G^+)$ is the worst, indicating the manipulation captures a mechanism-level effect rather than a model-idiosyncratic artefact.

\paragraph{Helpful-side companion factorial.} We mirror the same $2{\times}2$ design with persistent helpful memory on Qwen3.5-27B (Table~\ref{tab:mtb_factorial2_helpful}). The cell-level gains are an order of magnitude smaller than the corresponding conflicting-side damages, and $E$ and $G$ are roughly additive ($\approx +3$\pp{} each, max $+6.2$\pp{} at $(E^+,G^+)$). Comparing tables, conflicting damage is gate-multiplicative ($-53.1$\pp{} at $(E^+,G^+)$ vs.\ $-3.1$\pp{} at $(E^-,G^-)$, an $\approx 17\times$ ratio), while helpful gain is gate-additive. This is the gate-chain prediction of the discussion: conflicting memory clears the early gates trivially and is bottlenecked only at the terminal lift, so closing any one gate removes most of the damage; helpful memory leaks at every gate, so the maximum gain is small to begin with and closing one gate makes proportionally less difference. We omit Gemma-4-26B-A4B from the helpful side because its per-cell helpful response is at the level of cell-to-cell noise, consistent with the small Gemma-4 helpful gains in Table~\ref{tab:memtrapbench}.

\begin{table}[h]
\caption{\textbf{Helpful-side $2{\times}2$ factorial on the \textsc{Decoy} diagnostic} (Qwen3.5-27B; 32 tasks per cell, paired $\Delta$ vs.\ a same-task no-memory baseline of $65.6\%$). Helpful gains are bounded ($\leq +6.2$\pp{} at every cell) and roughly additive in $E$ and $G$, in contrast to the gate-multiplicative conflicting-side damages of Table~\ref{tab:mtb_factorial2}.}
\label{tab:mtb_factorial2_helpful}
\centering
\small
\begin{tabular}{l l c c}
\toprule
$E$ & $G$ & Cell SR & $\Delta$ vs.\ no memory \\
\midrule
multi  & ungroundable & 65.6\%          & $\phantom{+}0.0$ \\
multi  & groundable   & 68.8\%          & $+3.1$ \\
single & ungroundable & 68.8\%          & $+3.1$ \\
single & groundable   & \textbf{71.9\%} & $\textbf{+6.2}$ \\
\midrule
$E$ main effect          & & \multicolumn{2}{c}{$+3.1$\pp} \\
$G$ main effect          & & \multicolumn{2}{c}{$+3.1$\pp} \\
$E{\times}G$ interaction & & \multicolumn{2}{c}{$\approx 0$} \\
\bottomrule
\end{tabular}
\end{table}

\subsection{Memory delivery channel}

\paragraph{Setup.} The main text reports the persistent-conflicting numbers in two delivery channels (system prompt vs.\ observation footer; Table~\ref{tab:footer_main}). Here we add the persistent-helpful side and report the full $\pm$pp summary across three models.

\paragraph{What the helpful side adds.} Conflicting damage is invariant to placement (within $\pm 0.5$\pp), so the trap is content-driven. Helpful gain, by contrast, attenuates by $4$--$16$\pp{} under footer injection, most severely on Qwen3.5-27B ($+30.7 \to +14.7$\pp, a $52\%$ reduction). The asymmetry between the two directions matches the gate chain: conflicting memory only needs to trigger a plausible wrong action, while helpful memory must additionally clear adoption and grounding gates that prominent placement helps satisfy~\citep{liu2024lost, hagstrom2025cub}.

\begin{table}[h]
\caption{\textbf{Footer vs.\ system-prompt injection on MemTrapBench (231 long-horizon tasks).} Conflicting damage is invariant to channel (within $\pm 0.5$\pp); helpful gain attenuates by $4$--$16$\pp.}
\label{tab:footer_ablation}
\centering
\small
\begin{tabular}{l c cc cc}
\toprule
& & \multicolumn{2}{c}{Conflicting $\Delta$ vs.\ no\_mem} & \multicolumn{2}{c}{Helpful $\Delta$ vs.\ no\_mem} \\
\cmidrule(lr){3-4}\cmidrule(lr){5-6}
Model & Baseline & System prompt & \textbf{Footer} & System prompt & \textbf{Footer} \\
\midrule
Qwen3.5-9B          & 10.8\% & $-$9.5\pp  & $-$10.0\pp & $+$25.5\pp & $+$19.0\pp \\
Qwen3.5-27B         & 22.5\% & $-$21.2\pp & $-$20.8\pp & $+$30.7\pp & $+$14.7\pp \\
Gemma-4-26B-A4B     & 12.1\% & $-$10.8\pp & $-$10.4\pp & $+$29.9\pp & $+$25.5\pp \\
\bottomrule
\end{tabular}\\[2pt]
{\footnotesize Conflicting location-invariance: max $|\Delta_{\text{sys}} - \Delta_{\text{footer}}| = 0.5$\pp. Helpful attenuation under footer: 9B $-6.5$\pp, 27B $-16.0$\pp, Gemma-4-26B-A4B $-4.4$\pp.}
\end{table}

\section{Boundary conditions}
\label{app:gemma3}
\label{app:crossfamily}
\label{app:miniwob}
\label{app:short_horizon}

This appendix asks where the trap fades. Two natural boundaries: a different model family (does the trap depend on the Qwen3.5/Gemma-4 architectures?) and a shorter horizon (does it depend on having a long trajectory in which to propagate?).

\subsection{Cross-family validation: Gemma-3}

\paragraph{Question.} Does the compliance-floor mechanism survive when we move to a different family of open-weight models?

\paragraph{Result.} On 611 common valid WebArena tasks with online-retrieved memory, Gemma-3 shows the opposite scaling direction from Qwen3.5: the larger Gemma-3 model is \emph{more} hurt by memory, while the larger Qwen model benefits (Table~\ref{tab:crossfamily}). This is consistent with the conditional mechanism: vulnerability depends on the interaction between baseline capability, compliance rate, and the post-compliance success floor. Larger models are not automatically more vulnerable in every setting; they become more vulnerable when they start from a higher baseline and still comply at a similar rate, so the same low floor creates a larger absolute drop. Architecture and instruction-tuning calibration~\citep{ouyang2022instructgpt, bai2022constitutional} can shift the baseline and the floor without removing the mechanism.

\begin{table}[h]
  \caption{\textbf{Cross-family early injection on 611 common valid WebArena tasks.} Gemma-3 shows the opposite scaling direction from Qwen3.5: the larger Gemma-3 model is \emph{more} hurt by memory, while the larger Qwen model benefits. At 27B parameters the two families diverge by 5.1\pp.}
  \label{tab:crossfamily}
  \centering
  \small
  \begin{tabular}{llcccc}
    \toprule
    Family & Model & Params & B1 SR & M3 SR & $\Delta$ \\
    \midrule
    \multirow{2}{*}{Qwen3.5} & 9B & 9B & 21.4\% & 19.4\% & $-$2.0\pp \\
     & 27B & 27B & 25.1\% & 26.8\% & $+$1.7\pp \\
    \midrule
    \multirow{3}{*}{Gemma-3} & 4B-it & 4B & 11.4\% & 12.4\% & $+$1.0\pp \\
     & 12B-it & 12B & 16.2\% & 16.0\% & $-$0.3\pp \\
     & 27B-it & 27B & 19.9\% & 16.5\% & $-$3.4\pp$^{***}$ \\
    \bottomrule
    \multicolumn{6}{l}{\footnotesize $^{***}$ $p = 0.001$ (sign-flip permutation test, 10{,}000 iterations).}
  \end{tabular}
\end{table}

\subsection{Short-horizon control: MiniWoB++}

\paragraph{Question.} Does the trap appear when the trajectory is too short for propagation and recovery to operate?

\paragraph{Result.} Largely no. On MiniWoB++~\citep{shi2017miniwob} (12 tasks $\times$ 7 temporal conditions $\times$ 3 random seeds, median trajectory $\approx 2$ steps), persistent-conflicting drops are only $-2.9$ to $0$\pp{} and persistent-helpful is $-8.3$ to $0$\pp{} (Table~\ref{tab:miniwob}). Two reasons make this expected under E-P-R. First, baselines of $36$--$40\%$ on $1$--$3$ step tasks leave little room for the success-lift gate to operate. Second, with a median of $2$ steps, persistent, late, and early injection are nearly the same event, so propagation and recovery are not differentiable. Long-horizon interaction (median $8$--$15$ steps) is therefore a precondition for our findings; this matches the regime where coverage benchmarks such as WebVoyager~\citep{he2024webvoyager} and TheAgentCompany~\citep{xu2024theagentcompany} also report the largest absolute gaps between models.

\begin{table}[h]
  \caption{\textbf{MiniWoB++ short-horizon temporal injection.} Unlike WebArena and MemTrapBench, persistent-conflicting memory produces only $-2.9$ to $0$\pp{} drops; persistent-helpful sometimes \emph{hurts}. The compliance trap does not replicate at short horizons.}
  \label{tab:miniwob}
  \centering
  \small
  \begin{tabular}{lcccccc}
    \toprule
    Model & Base & Early H & Early C & Pers.\ H & Pers.\ C & Late H \\
    \midrule
    Qwen3.5-9B          & 38.9\% & $-$5.6 & +2.8 & $-$8.3 & $+$0.0 & $-$8.3 \\
    Qwen3.5-27B         & 40.0\% & $+$0.0 & +1.7 & $-$3.9 & $-$2.9 & $-$6.7 \\
    Gemma-4-26B-A4B     & 36.1\% & $+$0.0 & $+$0.0 & $+$0.0 & +3.9 & +2.8 \\
    \bottomrule
  \end{tabular}
\end{table}

\section{Retry-on-fail ablations}
\label{app:retry_ablation}

The main text reports retry-on-fail as a minimal demonstration that conditioning memory injection on observed first-attempt failure outperforms uniform injection. This appendix gives the two ablations that support that claim: a head-to-head comparison against fixed-schedule policies on WebArena, and a content-vs-sampling decomposition on MemTrapBench.

\paragraph{Lightweight policies on WebArena 77.} Table~\ref{tab:policy_comparison} compares fixed-schedule helpful injection, groundability gating, and retry-on-fail against the schedule oracle. Always-persistent helpful captures only $50\%$ and $17\%$ of the oracle gap on Qwen3.5-9B and Qwen3.5-27B; URL-keyword groundability gating does not solve the problem and even hurts the larger model. Retry-on-fail closes $50\%$ of the gap on Qwen3.5-9B and $133\%$ on Qwen3.5-27B; the latter overshoot is consistent with small-sample variation, and the MemTrapBench ablation below shows the same model at $96\%$ closure on a $\sim 3{\times}$-larger task pool.

\begin{table}[h]
\caption{\textbf{Lightweight injection policies on WebArena 77.}
Each cell reports absolute success with $\Delta$ versus no-memory baseline and oracle closure. Retry-on-fail uses only the binary first-attempt outcome.}
\label{tab:policy_comparison}
\centering
\small
\begin{tabular}{lccl}
\toprule
Policy & Qwen3.5-9B & Qwen3.5-27B & Signal used \\
\midrule
No memory (baseline)            & 29.9\% (---)                    & 40.3\% (---)                     & --- \\
Always persistent helpful       & 35.1\% ($+$5.2 / 50\%)          & 41.6\% ($+$1.3 / 17\%)           & none \\
Always late helpful$^{\dagger}$ & 30.2\% ($+$0.3 / 3\%)           & 42.7\% ($+$2.4 / 32\%)           & trajectory step \\
Groundability gating            & 31.2\% ($+$1.3 / 13\%)          & 39.0\% ($-$1.3 / $-$17\%)        & URL / keyword match \\
\midrule
\textbf{Retry-on-fail}          & \textbf{35.1\% ($+$5.2 / 50\%)} & \textbf{50.6\% ($+$10.4 / 133\%)} & first-attempt outcome \\
\midrule
Schedule oracle                 & 40.3\% ($+$10.4 / 100\%)        & 48.1\% ($+$7.8 / 100\%)          & best fixed schedule per task \\
\bottomrule
\end{tabular}\\[2pt]
{\footnotesize Closure is the policy's $\Delta$ divided by the paired schedule-oracle $\Delta$. ${}^{\dagger}$Computed on the late-helpful-valid subset ($n=70$--$76$); absolute deltas differ by at most $\pm 3$\pp{} from Figure~\ref{fig:schedule_bars}.}
\end{table}

\paragraph{Retry-helpful vs.\ retry-nomem on MemTrapBench.} Does the gain come from the second attempt's sampling diversity, or from the helpful memory it now sees? We separate the two by running \textbf{retry-nomem} (Attempt 1 no-memory; on failure Attempt 2 also no-memory) and \textbf{retry-helpful} (Attempt 1 no-memory; on failure Attempt 2 with persistent helpful memory) on all 231 main-set MemTrapBench tasks across three models (Table \ref{tab:retry_ablation_mtb}). Both arms use the same one-bit failure signal, so the difference retry-helpful $-$ retry-nomem isolates the marginal benefit of memory content. Retry without memory contributes only $+2.6$ to $+5.6$\pp{} (sampling alone), while retry with helpful memory contributes $+19.7$ to $+31.2$\pp{}; the gap is $+14$ to $+28$\pp. On Qwen3.5-27B and Gemma-4-26B-A4B retry-helpful effectively closes the schedule-oracle gap (96\% and 101\%); Qwen3.5-9B closes 76\%, consistent with the fourth-gate (success-lift) bottleneck of the discussion: smaller models extract less from the same helpful content.

\begin{table}[h]
\caption{\textbf{Retry ablations on MemTrapBench (231 long-horizon tasks).} retry-nomem alone contributes only $+$2.6--5.6\pp{} (sampling variation); retry-helpful adds $+$19.7--31.2\pp. The $+$14--28\pp{} gap attributes the recovery to memory content rather than to second-attempt diversity. Recovery rate (\% of retry-triggered tasks that succeed on Attempt 2) is 35--40\% with helpful memory vs.\ 3\% without, a more than ten-fold lift.}
\label{tab:retry_ablation_mtb}
\centering
\small
\begin{tabular}{l ccc ccc}
\toprule
& \multicolumn{3}{c}{Overall SR ($\Delta$ vs.\ no\_mem)} & \multicolumn{3}{c}{Recovery rate (Attempt 2 $\mid$ Att.\,1 failed)} \\
\cmidrule(lr){2-4}\cmidrule(lr){5-7}
Model & no\_mem & retry-nomem & retry-helpful & triggered & nomem & helpful \\
\midrule
Qwen3.5-9B          & 10.8\% & 12.8\% ($+$2.0)  & 30.5\% ($+$19.7) & 89\% & \phantom{1}2\% & 22\% \\
Qwen3.5-27B         & 22.5\% & 25.1\% ($+$2.6)  & 53.7\% ($+$31.2) & 78\% & \phantom{1}3\% & 40\% \\
Gemma-4-26B-A4B     & 12.1\% & 14.8\% ($+$2.7)  & 43.2\% ($+$31.1) & 88\% & \phantom{1}3\% & 35\% \\
\bottomrule
\end{tabular}\\[2pt]
{\footnotesize retry-helpful $-$ retry-nomem gap: 9B $+$17.7\pp, 27B $+$28.6\pp, Gemma-4-26B-A4B $+$28.4\pp. Oracle gap closure (retry-helpful $\Delta$ / oracle $\Delta$): 9B 76\%, 27B 96\%, Gemma-4-26B-A4B 101\%. Denominators: 27B $n=$231; 9B $n=$211--213; Gemma-4-26B-A4B $n=$229.}
\end{table}

\paragraph{Why this is a lower bound.} Retry-on-fail uses the smallest possible amount of state information (whether the first attempt failed) to route memory better than always injecting. Better policies should predict memory usefulness before failure rather than after it; this is conceptually adjacent to Self-Refine~\citep{madaan2023selfrefine} but applied to memory selection rather than output revision. The Qwen3.5-27B WebArena overshoot ($+2.5$\pp{} over the schedule oracle, Table~\ref{tab:policy_comparison}) reads cleanly under these ablations: on the $77$-task subset, retry-helpful benefits both from memory content and from the favorable intersection of first-failure sampling with retry; on the $\sim 3{\times}$-larger MemTrapBench, the same model closes 96\% of the gap rather than overshooting it, confirming that the WebArena surplus reflects small-sample variation rather than a stable property.


\newpage

\end{document}